\documentclass[sigconf]{acmart}
\AtBeginDocument{%
  }

\copyrightyear{2025}
\acmYear{2025}
\setcopyright{cc}
\setcctype{by}
\acmConference[KDD '25]{Proceedings of the 31st ACM SIGKDD Conference on Knowledge Discovery and Data Mining V.2}{August 3--7, 2025}{Toronto, ON, Canada}
\acmBooktitle{Proceedings of the 31st ACM SIGKDD Conference on Knowledge Discovery and Data Mining V.2 (KDD '25), August 3--7, 2025, Toronto, ON, Canada}
\acmDOI{10.1145/3711896.3737434}
\acmISBN{979-8-4007-1454-2/2025/08}

\usepackage{hyperref}
\usepackage{url}
\usepackage{booktabs}       
\usepackage{amsfonts}       
\usepackage{nicefrac}       
\usepackage{xcolor}         
\usepackage{tabularx}

\usepackage{amsmath}
\usepackage{caption}
\usepackage{amsfonts}
\usepackage{setspace}
\usepackage{multirow}
\usepackage{makecell}
\usepackage{subfigure}
\usepackage{color}
\usepackage{xspace}
\usepackage{amsmath}
\usepackage{adjustbox}
\usepackage{booktabs}
\usepackage{soul}
\usepackage{float}
\usepackage{textcomp}
\usepackage{enumitem}
\usepackage{algorithm}
\usepackage{algpseudocode}
\usepackage{tabularray}
\usepackage{wrapfig}
\usepackage{makecell}
\usepackage{cleveref}
\usepackage{multicol}
\usepackage{longtable}

\definecolor{firstcolor}{HTML}{1cc61c}
\definecolor{secondcolor}{HTML}{cb0707}
\newcommand\colorup[1]{\textcolor{firstcolor}{\textbf{#1}}}
\newcommand\colordown[1]{\textcolor{secondcolor}{\textbf{#1}}}

\DeclareMathOperator*{\argminA}{arg\,min}

\begin{document}

\title[Towards Understanding Link Prediction Shift]{Towards Understanding Link Predictor \\ Generalizability Under Distribution Shifts}

\author{Jay Revolinsky}
\authornote{Both authors contributed equally to this research.}
\affiliation{%
  \institution{Michigan State University}
  \city{East Lansing}
  \state{}
  \country{USA}
}
\email{revolins@msu.edu}

\author{Harry Shomer}
\authornotemark[1]
\affiliation{%
  \institution{Michigan State University}
  \city{East Lansing}
  \state{}
  \country{USA}}
\email{shomerha@msu.edu}

\author{Jiliang Tang}
\affiliation{%
  \institution{Michigan State University}
  \city{East Lansing}
  \state{}
  \country{USA}
}
\email{tangjili@msu.edu}

\renewcommand{\shortauthors}{Revolinsky et al.}

\begin{abstract}
  State-of-the-art link prediction (LP) models demonstrate impressive benchmark results. However, popular benchmark datasets often assume that training, validation, and testing samples are representative of the overall dataset distribution. In real-world situations, this assumption is often incorrect; uncontrolled factors lead new dataset samples to come from a different distribution than training samples. Additionally, the majority of recent work with graph dataset shift focuses on node- and graph-level tasks, largely ignoring link-level tasks. To bridge this gap, we introduce a novel splitting strategy, known as LPShift, which utilizes structural properties to induce a controlled distribution shift. We verify LPShift's effect through empirical evaluation of SOTA LP models on 16 LPShift variants of original dataset splits, with results indicating drastic changes to model performance. Additional experiments demonstrate graph structure has a strong influence on the success of current generalization methods. The source code is available here \href{https://github.com/revolins/LPShift}{https://github.com/revolins/LPShift}.
\end{abstract}




\maketitle

\section{Introduction} \label{sec:intro}
  
Link Prediction (LP) is concerned with predicting unseen links (i.e., edges) between two nodes in a graph \cite{liben2003link}. The task has a wide variety of applications including: recommender systems, \cite{fan2019graph}, knowledge graph completion \cite{lin2015learning}, protein-interaction \cite{kovacs2019network}, and drug discovery \cite{abbas2021application}. Traditionally, LP was performed using heuristics that model the pairwise interaction between two nodes~\cite{newman2001clustering, zhou2009predicting, adamic2003friends}.
The success of Graph Neural Networks (GNNs)~\cite{kipf2017semi} has prompted their usage in LP~\cite{kipf2016variational, zhang2018link}. However, GNNs are unable to fully-capture representations for node pairs~\cite{zhang2021labeling, srinivasan2019equivalence}. To combat this problem, recent methods (i.e., GNN4LP) empower GNNs with additional information to capture pairwise interactions between nodes~\cite{shomer2024lpformer, chamberlain2022graph, zhang2018link, wang2023neural} and demonstrate tremendous ability to model LP on real-world datasets~\cite{hu2020open}.

While recent methods have shown promise, current benchmarks \cite{hu2020open} typically assume that the training and evaluation data is {\it drawn from the same structural distribution}. This assumption collapses in real-world scenarios, where the structural feature (i.e., covariate) distribution may shift from training to evaluation. Therefore, it's often necessary for models to generalize to samples whose newly-introduced feature distribution differs from the training dataset~\cite{finedistshift, wildtime,yao2022sela, bevilacqua2021size}.

Furthermore, while numerous methods work to account for distribution shifts within graph machine learning~\cite{li2022out}, there remains little work doing so for LP. Specifically, we observe that {\bf (1) No LP Benchmark Datasets}: Current graph benchmark datasets designed with a quantifiable distribution shift are focused solely on the node and graph tasks~\cite{zhou2022ood, ding2021closer}, {\it with no datasets available for LP}. {\bf (2) Absence of Foundational Work}: There is limited existing work for distribution shifts relevant to LP \citep{zhang2022dynamic, dong2022fakeedge}. Current methods are primarily focused on detecting and alleviating anomalies within node- and graph-level tasks~\cite{jin2022empowering, bevilacqua2021size, gao2023alleviating, wu2024graph, guo2023data, wu2023energy, sui2024unleashing, li2022out}. Additionally, few methods are designed for aiding LP generalization in any setting \cite{dong2022fakeedge, zhao2022learning, zhang2022dynamic, zhou2022ood}. Also, other LP generalization methods which are theorized to improve performance in shifted scenarios remain crucially untested \cite{singh2021edge, wang2023topological}. 

To tackle these problems, this work proposes the following contributions:
{\bf (1)} \textit{Creating Datasets with Meaningful Distribution Shifts}. LP requires pairwise structural considerations \cite{liben2003link, mao2024demystifying}. Additionally, when considering realistic settings \cite{li2024evaluating} or distribution shift \cite{zhu2024pitfalls}, GNN4LP models are sensitive to structual changes within datasets relative to graph \cite{PASGraph, yuan2021largescale} and node classification \cite{shi2023label, zhao2023learning} models. To quantify this sensitivity and better understand distribution shifts, we use key structural LP heuristics to split the links into train/validation/test splits via LPShift. By applying LPShift to generate dataset splits, we induce shifts in the underlying distribution of the links, thereby affecting link formation~\cite{mao2024demystifying} and downstream performance. This elevates preexisting LP benchmarks by allowing users to control the extent with which their dataset undergoes distribution shift while retaining relevance to LP. Much like how prior benchmarks induce substructure~\cite{gui2022good, koh2021wilds} and drug scaffold~\cite{ji2022drugood} shift to provide relevant challenges within graph and node classification. Theoretical justification is provided in Section~\ref{sec:why_lpshift} and empirical results are provided in Section~\ref{sec:experiments}. {\bf (2)} \textit{Benchmarking Current LP Methods}. GNN4LP models demonstrate a strong sensitivity to LPShift when generalizing to the synthetic dataset splits. Despite the existence of LP generalization methods, such as FakeEdge \cite{dong2022fakeedge} and Edge Proposal Sets \cite{singh2021edge}, there remains little work benchmarking link-prediction models under distribution shifts \cite{ding2021closer, dong2022fakeedge, zhu2024pitfalls}. This lack of benchmarking contributes to a gap in understanding, impeding the capabilities of LP models to generalize in real-world scenarios. This work quantifies the performance of SOTA LP models under 16 unique LPShift scenarios and provides analysis as a foundation for improving LP model generalization. We further quantify the effects of LP and graph-specific generalization methods, finding that they also struggle to generalize with differing structural shifts.

The remainder of this paper is structured as follows. In Section~\ref{sec:related_work}, we provide background on the heuristics, models, and generalization methods used in LP. In Sections~\ref{sec:create_datasets} and~\ref{sec:why_lpshift}, we detail how structural heuristics, theoretical perspectives, and HeaRT~\cite{li2024evaluating} relate to our proposed splitting strategy. After which, we formally introduce LPShift. Lastly, in Section~\ref{sec:experiments}, we benchmark a selection of LP models and generalization methods on LPShift, followed by analysis to understand the effects of this new strategy.

\section{Related Work} \label{sec:related_work}

\noindent {\bf LP Heuristics}: Classically, neighborhood heuristics, which measure characteristics between source and target edges, functioned as the primary means of predicting links. These heuristics show limited effectiveness with a relatively-high variability in results, largely due to the complicated irregularity within graph datasets; which only grows worse with larger datasets \cite{liben2003link}. Regardless of this, state-of-the-art GNN4LP models have integrated these neighborhood heuristics into neural architectures to elevate link prediction capabilities \cite{wang2023neural, chamberlain2022graph}. 

For a given heuristic function, $u$ and $v$ represent the source and target nodes in a potential link, $(u, v)$. $\mathcal{N}(v)$ is the set of all edges, or neighbors, connected to node $v$. $f(v_{i,i + 1}, u)$ is a function that considers all paths of length $i$ that start at $v$ and connect to $u$. The three tested heuristics are as follows:

{\it Common Neighbors}~\cite{newman2001clustering}: The number of neighbors shared by two nodes $u$ and $v$, 
\begin{equation}
    \text{CN}(u,v) = | \mathcal{N}(u) \cap \mathcal{N}(v) |.
\end{equation}
{\it Preferential Attachment}~\cite{liben2003link}: The product of the number of neighbors (i.e., the degree) for nodes $u$ and $v$,
\begin{equation}
    \text{PA}(u,v) = | \mathcal{N}(u) | \times | \mathcal{N}(v) |.
\end{equation}

{\it Shortest Path Length}~\cite{liben2003link}: The path between $u$ and $v$ which considers the smallest possible number of nodes, denoted as length $n$, 
\begin{equation}
    \text{SP}(u,v) = \argminA_{\Sigma}( \Sigma^{n - 1}_{i = 1} f(v_{i,i + 1}, u)) .
\end{equation}

\noindent {\bf GNNs for Link Prediction (GNN4LP)}: LP's current SOTA methods rely on GNNs for a given model's backbone. The most common choice is the Graph Convolutional Network (GCN)~\cite{kipf2017semi}, integrating a simplified convolution operator to consider a node's multi-hop neighborhood. The final score (i.e., probability) of a link existing considers the representation between both nodes of interest. However, \cite{zhang2021labeling} show that such methods aren't suitably expressive for LP, as they ignore vital pairwise information that exists between both nodes. To account for this, SEAL~\cite{zhang2018link} conditions the message passing on both nodes in the target link by applying a node-labelling trick to the enclosed k-hop neighborhood. They demonstrate that this can result in a suitably expressive GNN for LP. NBFNet~\cite{zhu2021neural} conditions the message passing on a single node in the target link by parameterizing the generalized Bellman-Ford algorithm. In practice, it's been shown that conditional message passing is prohibitively expensive to run on many LP datasets~\cite{chamberlain2022graph}. Instead, recent methods take the standard GNN representations and an additional pairwise encoding into their scoring function for link prediction. For the pairwise encoding, Neo-GNN~\cite{yun2021neo} considers the higher-order overlap between neighborhoods. BUDDY~\cite{chamberlain2022graph} estimates subgraph counts via sketching to infer information surrounding a target link. Neural Common-Neighbors with Completion (NCNC)~\cite{wang2023neural} encodes the enclosed 1-hop neighborhood of both nodes. Lastly, LPFormer~\cite{shomer2024lpformer} adapts a transformer to learn the pairwise information between two nodes.

\noindent {\bf Generalization in Link Prediction}: Generalization methods for LP rely on a mix of link and node features in order to improve LP model performance. DropEdge~\cite{rong2020dropedge} randomly removes edges with increasing probability from the training adjacency matrix, allowing for different views of the graph. Edge Proposal Sets (EPS)~\cite{singh2021edge} considers two models -- a filter and rank model. The filter model is used to augment the graph with top-k relevant common neighbors, while the rank method scores the final prediction. \cite{wang2023topological} built Topological Concentration (TC), which considers the overlap in subgraph features for a given node with each connected neighbor, correlating well with LP performance for individual links. To improve the performance of links with a low TC, a re-weighting strategy applies more emphasis on links with a lower TC. Counter-Factual Link Prediction (CFLP)~\cite{zhao2022learning} conditions a pre-trained model with edges that contain information counter to the original adjacency matrix, allowing models to generalize on information not present in a provided dataset.
\section{Benchmark Dataset Construction} \label{sec:create_datasets}

\begin{figure*}[ht]
     \begin{center}
      \centerline{
         {\subfigure[{\large (a) CN}]
         {\includegraphics[width=0.22\linewidth]{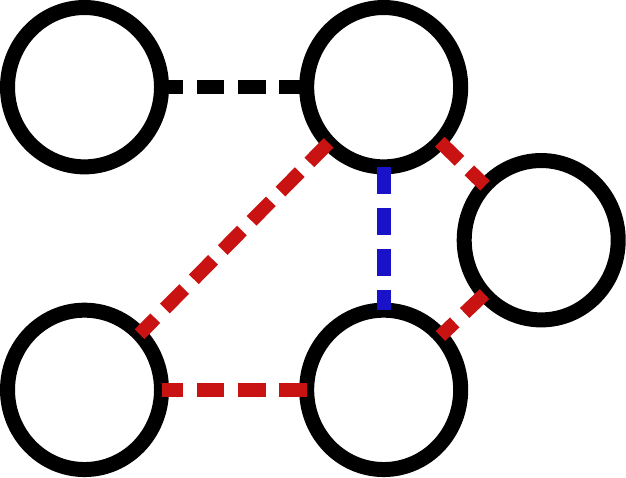} \label{fig:cn_split_ogb} } \hspace*{1.5em}}
        
         {\subfigure[{\large (b) PA}]
         {\includegraphics[width=0.28\linewidth]{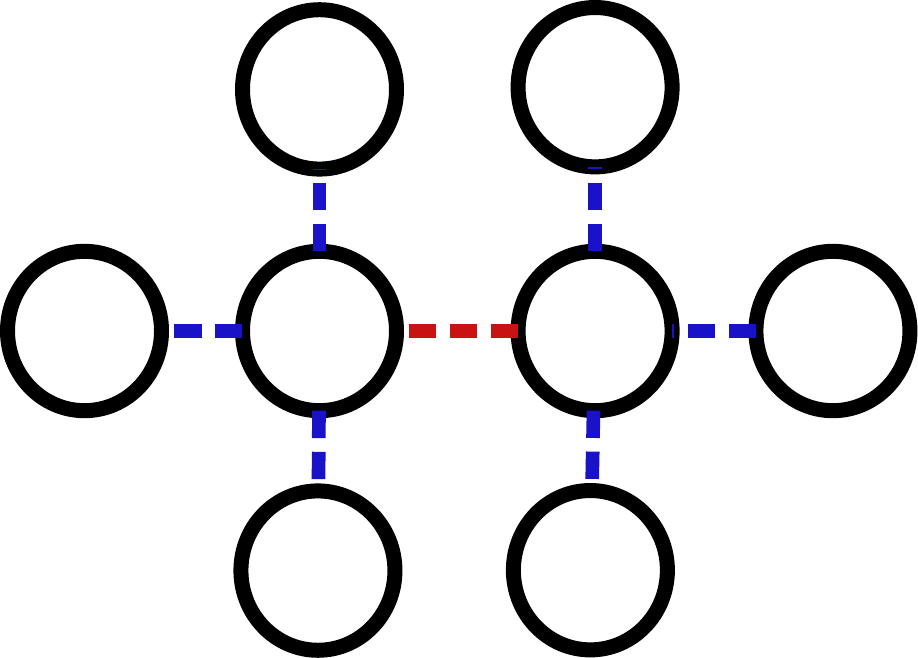} 
         \label{fig:pa_split_ogb}} \hspace*{1.5em}}
        
         {\subfigure[{\large (c) SP}]
         {\includegraphics[width=0.26\linewidth]{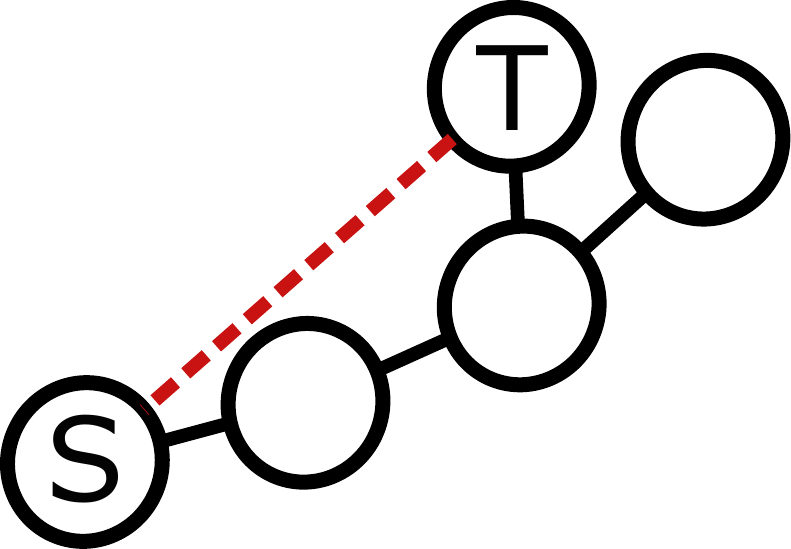} 
         \label{fig:sp_split_ogb} }}
     }
    
     \caption{An example of the three splitting strategies: (a) Common Neighbors, (b) Preferential-Attachment, (c) Shortest-Path. The dashed lines represent a cut on the source and target node that forms a given edge for our splitting strategy. The color of the line distinguishes the score assigned by the heuristic.}
     \Description{An example of the three splitting strategies: (a) Common Neighbors, (b) Preferential-Attachment, (c) Shortest-Path. The dashed lines represent a cut on the source and target node that forms a given edge for our splitting strategy. The color of the line distinguishes the score assigned by the heuristic.}

 \label{fig:split_ogb}
 \end{center}
 \end{figure*}

In this section, we explain how LPShift induces a shift in each dataset's structure; clarifying the importance of each structural measure and their application to splitting graph data.

\subsection{Types of Distribution Shifts}

We induce distribution shifts by splitting the links based on key structural properties affecting link formation and thereby LP. We consider three type of metrics: Local structural information, Global structural information, and Preferential Attachment. Recent work by~\cite{mao2023revisiting} has shown the importance of local and global structural information for LP. Furthermore, due to the scale-free nature of many real-world graphs and how it relates to link formation \cite{BAGraphs}, we also consider Preferential Attachment. A representative metric is then chosen for each of the three types, shown as follows:  

\noindent {\bf (1) Common Neighbors (CNs)}: CNs measure {\it local structural information} by considering only those nodes connected to the target and source nodes. A real-world case for CNs is whether you share mutual friends with a random person, thus determining if they are your ``friend-of-a-friend'' \cite{adamic2003friends}. CNs plays a large role in GNN4LP, given that NCNC \cite{wang2023neural} and EPS \cite{singh2021edge} integrate CNs into their framework and achieve SOTA performance. Furthermore, even on complex real-world datasets, CNs achieves competitive performance against more advanced neural models \cite{hu2020open}. To control for the effect of CNs on shifted performance, the relevant splits will consider thresholds which include more CNs.

\noindent {\bf (2) Shortest Path (SP)}: SP captures a graph's {\it global structural information}, thanks to the shortest-path between a given target and source node representing the most efficient path for reaching the target \cite{AIMABook}. The shift in global structure caused by splitting data with SP can induce a scenario where a model must learn how two dissimilar nodes form a link with one another \cite{evtushenko2021paradox}, which is comparable to the real-world scenario where two opponents choose to co-operate with one another \cite{schelling1978micromotives, granovetter1978threshold}.

\noindent {\bf (3) Preferential Attachment (PA)}: PA captures the {\it scale-free property} of larger graphs by multiplying the degrees between two given nodes \cite{BAGraphs}. When applied to graph generation, PA produces synthetic Barabasi-Albert (BA) graphs which retain the scale-free property to effectively simulate the formation of new links in real-world graphs, such as the World Wide Web \cite{BAGraphs, AlbertComNet2002}.  Similar to CNs, the relevant PA splits will consider thresholds that integrate higher PA values.

\subsection{Dataset Splitting Strategy} \label{sec:d_splt_strat_dc}
In the last subsection we described the different types of metrics to induce distribution shifts for LP. The metrics cover fundamental structural properties that influence the formation of new links. Given the functionality of: CN, SP, and PA within link prediction~\cite{liben2003link}, LPShift applies the three heuristics to induce distribution shifts which respectively target: local, global, and scale-free properties of the graph. Therefore, providing a meaningfully-difficult challenge to GNN4LP, as empirically-demonstrated in Section~\ref{sec:experiments}. We now describe how LPShift applies these heuristic measures to split the dataset into train/validation/test splits to induce such shifts.

\begin{figure*}[ht]
\centering
\includegraphics[width=0.99\linewidth]{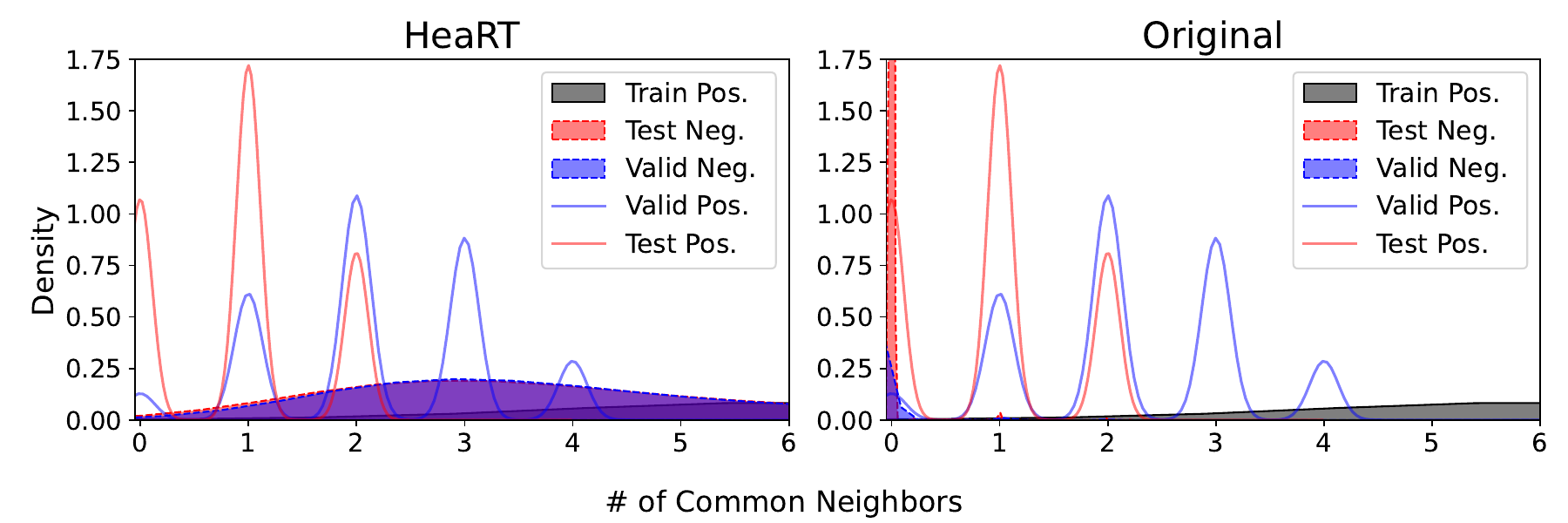}
\caption{The Common Neighbors distribution of the HeaRT versus the original testing and validation negative samples from LPShift's ogbl-collab 'CN - 5,3' dataset.}
\Description{The Common Neighbors distribution of the HeaRT testing and validation negative samples versus the original testing and validation negative samples compared against the positive samples drawn from LPShift's ogbl-collab 'CN - 5,3' dataset.}
\label{fig:why_heart_w_lpshift}
\end{figure*}

In order to build datasets with structural shift, we apply a given neighborhood heuristic to score each link. This score is then compared to a threshold ($i_{train}, i_{valid})$ to categorize a link within a different sample distribution. As denoted in Alg.~\ref{alg:dsplit}, the heuristic score of the link $(u, v)$ is $h(u, v)$. The link falls into: training when $h(u, v) < i_{train}$, validation when $i_{train} < h(u, v) \leq i_{valid}$, and testing when $h(u, v) > i_{valid}$. The new training graph is constructed from the original OGB dataset \cite{hu2020open}. Validation and testing samples are removed from the new training graph to prevent test-leakage and limited to 100-thousand edges maximum. 

\begin{algorithm}[h]
\caption{Dataset Splitting Strategy (LPShift) }
\begin{algorithmic}[1]

\Require
    \Statex $G = $ Initial Graph, 
    \Statex $\Psi(.,.) = $ Heuristic function
    \Statex $i_{train}$, $i_{valid} = $ Heuristic score thresholds
    \Statex $Train, Valid, Test = \emptyset, \emptyset, \emptyset$
\newline
\While{edge, $(u,v)$ not visited in $G$}
    \State $\Psi(u,v) = h(u,v)$ \Comment{ Score edge with heuristic}
    \If{$h(u,v) \leq i_{train}$} 
        \State $Train \gets (u,v)$
    \ElsIf{$h(u,v) > i_{train} \text{ and } h(u,v) \leq i_{valid}$} 
        \State $Valid \gets (u,v)$
    \Else \Comment{ $h(u,v) > i_{valid}$, Test Split} 
        \State $Test \gets (u,v)$
    \EndIf
\EndWhile
\State \Return $Train, Valid, Test$ \Comment{Return Final Splits}
\end{algorithmic}
\label{alg:dsplit}
\end{algorithm}

With Figure~\ref{fig:split_ogb}, we provide a small example of how splits are produced by our proposed splitting strategy. Specifically, Figure~\ref{fig:cn_split_ogb}(a) demonstrates an outcome of the CN split labelled ``CN - 1,2'' where sampled edges pulled from the: black-dotted line = training (no CNs), red-dotted line = validation, (1 CN), and blue-dotted line = testing ($\geq$2 CNs). Figure~\ref{fig:split_ogb}(b) follows a similar high-level process, focusing on node-degree instead of CNs. As such, the validation (red) cut splits nodes with degree 3 and the testing (blue) cut splits nodes with degree 1. Figure~\ref{fig:split_ogb}(c) details Shortest-Path (SP), where the red cut works on a path of length 3 and is categorized by taking SP as the denominator of the ratio, $\frac{1}{SP}$. 

To test how different LPShift thresholds impact performance, we adjust the $i_{train}$ and $i_{valid}$ thresholds to produce 3 varied CN and PA splits; as well as 2 varied SP splits. The variations in split thresholds were chosen based on two conditions. \textbf{1)} Structural information within splits varies due to user-defined thresholds. In the "Forward" scenario (i.e. CN - 1,2), splits are given increasingly more structural information between training to validation, giving the model an easier time generalizing under testing. For the "Backward" scenario (i.e, CN - 2,1), stricter thresholds mean less structural information between validation to testing; making generalization difficult. \textbf{2)} The final dataset split contains a sufficient number of samples. Each LPShift split requires enough split samples to allow model generalization. Given the limited number of SP split samples, the SP splits were limited to 2 variants. 

\subsection{Why use HeaRT with LPShift?} 

\label{sec:heart_with_lpshift} It is common practice for most link prediction baselines to randomly-sample all negatives samples. However, random sampling is shown to be unrealistic, given that it is trivially-easy for a GNN4LP model trained via supervised learning to distinguish positive and negative samples based solely on the sample's structure~\cite{li2024evaluating}. We demonstrate an example of this effect in the right subplot of Figure~\ref{fig:why_heart_w_lpshift}, where the original setting samples negative at random to overwhelmingly contain 0 Common Neighbors. The samples produced by the original setting can be especially misleading, GNN4LP models could achieve high performance due to this overwhelming density of negative samples with 0 CNs versus positive samples with $>$0 CNs. Figure~\ref{fig:why_heart_w_lpshift}'s left subplot shows how HeaRT corrects this problem by sampling valid and test negatives whose distribution of CNs overlap with their respective positive samples (shown in red and blue), as well as the distribution of positive training samples (shown in black on both subplots). Therefore, HeaRT sampling avoids a critical pitfall within the original setting, where GNN4LP models could learn the trivial dynamic where \textit{any} sample with more than 0 CNs as a positive edge and any sample with 0 CNs as a negative edge. 

\section{Why LPShift? A Theoretical Perspective} \label{sec:why_lpshift} Section~\ref{sec:create_datasets} details how heuristic choice and HeaRT influence LPShift's ability to alter dataset structure. We now consider how GNNs, as neural backbones for GNN4LP, rely on assumptions about the dataset's underlying sample distributions to learn representations for effective link prediction. From these assumptions, GNN4LP capture structural heuristics associated with link formation, such as LPFormer estimating Katz-Index~\cite{shomer2024lpformer}. LPShift undermines these assumptions through direct manipulation of the dataset sample distribution, which is quantifiable within probability theory. Under which, we treat a link predictor as a function to learn a probability measure over a pairwise embedding. This learned probability measure, coupled with the obtained embedding, determines the ability of GNN4LP to quantify whether link formation occurs or not. We provide a theoretical perspective on this effect below: 

\paragraph{Notation:}  We denote the node feature space as \(\textbf{X} \in \mathbb{R}^{N \times F}\), where \(F\) is the number of node features, and the adjacency matrix representing a given graph dataset as, \(\textbf{A} \in \{0,1\}^{N \times N} \).

\begin{definition} \label{def:prob_defi}
Let \(\mathcal{P(\textbf{H})}\) represent a learned probability measure over \(\textbf{H}=GNN(\textbf{A,X})\). Given a sufficiently-large and i.i.d. dataset, it follows that an empirical measure on a dataset subset estimated by a GNN, \(\mathcal{\hat{P}\textbf{(H)}}\) over a subset of \(\textbf{A}\) will capture the underlying distribution \(\mathcal{P(\textbf{H})}\) for any dataset partition. 
\end{definition}


\begin{theorem} \label{def:psi_theor}
Let \({\Psi(u,v)}\) represent a function to capture pairwise-dependencies between a given edge, $(u, v)$, and subsequently partition a subset of edges from \((\textbf{A,X})\) by their structure, as determined by the output of \({\Psi(u,v)}\).
Given this subset partitioned by \({\Psi(u,v)}\), it follows that the newly-partitioned samples violate the i.i.d. assumptions from Definition~\ref{def:prob_defi}. Therefore, an empirical measure estimated by a GNN, \(\mathcal{\hat{P}(\textbf{H})}\) over dataset subsets partitioned by \(\Psi(u,v)\) will not estimate one another, i.e. $\mathcal{\hat{P}}^{Train}(\textbf{H}) \neq \mathcal{\hat{P}}^{Test}(\textbf{H})$.
\end{theorem}

\begin{proof}
Let \(\Psi(u,v)\) be a partition function over the dataset, $(\textbf{A, X})$ which defines partitioned subsets based on pairwise structures captured by $\textbf{A}$. Given the i.i.d. assumption about dataset samples, it follows that for distinct partitions $i \neq j$ and $\textbf{H} = GNN(\textbf{A,X})$, it holds that \(\mathcal{P}^{i}_{\Psi}(\textbf{H}) \neq \mathcal{P}^{j}_{\Psi}(\textbf{H})\). It then follows that \(\mathcal{\hat{P}}^{i}_{\Psi}(\textbf{H}) \sim \mathcal{P}^{i}_{\Psi}(\textbf{H})\) and \(\mathcal{\hat{P}}^{j}_{\Psi}(\textbf{H}) \sim \mathcal{P}^{j}_{\Psi}(\textbf{H})\). Therefore, \(\mathcal{\hat{P}}^{i}_{\Psi}(\textbf{H}) \neq \mathcal{\hat{P}}^{j}_{\Psi}(\textbf{H}) \equiv \mathcal{\hat{P}}^{Train}(\textbf{H}) \neq \mathcal{\hat{P}}^{Test}(\textbf{H})\).
\end{proof}

Indeed, when the i.i.d assumption across dataset samples holds, supervised training of neural LP models with GNN backbones enables message-passing across training samples. This further allows GNN4LP to observe and subsequently model link formation~\cite{bronstein2021geometric, zevcevic2021relating}, contributing to high downstream performance. However, Theorem~\ref{def:psi_theor} conditions link formation dynamics within a given graph dataset, meaning that the necessary i.i.d. subsets for learning and then achieving generalized performance are obscured. In effect, the GNN4LP model gets a limited view on pairwise information contained within the dataset. As such, LPShift allows direct testing of the hypothesis: \textit{Can GNN4LP model structural heuristics for effective downstream performance in scenarios with limited structural information?} We verify this effect empirically in Section~\ref{sec:experiments}.
\section{Experiments} \label{sec:experiments}

To bridge the gap for GNN4LP generalizing under  distribution shifts, this work addresses the following questions: {\bf (RQ1)} Is the distribution shift induced by LPShift significant? {\bf (RQ2)} Can SOTA GNN4LPs generalize under our proposed distribution shifts? {\bf (RQ3)} Can current generalization methods boost the performance of GNN models? {\bf (RQ4)} What components of the proposed distribution shift are affecting the LP model's performance? 

\subsection{Experimental Setup} \label{sec:experimental_setup}

{\bf Datasets}: We consider 16 ``Forward'' and ``Backward'' LPShift splits for the following 3 OGB datasets~\cite{hu2020open}: ppa, ddi, and collab, for a total of 48 tested splits. The resulting datasets represent tasks in three separate domains and three shifted scenarios, allowing a comprehensive study of LP generalization under distribution shift. For all datasets, we create multiple splits corresponding to each structural property, as detailed in Section~\ref{sec:d_splt_strat_dc}. For the ``Forward'' split, denoted as $(X,Y,Z)$, an increase in $Y$ and $Z$ indicates more structural information available to the training adjacency matrix. The ``Backward'' split swaps the training and testing splits from their counterpart in the ``Forward'' split, resulting in the training adjacency matrix losing access to structural information as $X$ and $Y$ increase.

{\bf GNN4LP Methods}: We test multiple SOTA GNN4LP methods including: NCNC \cite{wang2023neural}, BUDDY \cite{chamberlain2022graph}, LPFormer \cite{shomer2024lpformer}, SEAL \cite{zhang2021labeling} and Neo-GNN \cite{yun2021neo}. We further consider GCN~\cite{kipf2017semi} as a simpler GNN baseline, along with the Resource Allocation (RA)~\cite{zhou2009predicting} heuristic. All models were selected based on their benchmark performance with the original OGB datsets~\cite{hu2020open} and their architectural differences detailed in Section~\ref{sec:related_work}.

{\bf Graph-Specific Generalization Methods}: We also test the performance of BUDDY with multiple generalization techniques. This includes DropEdge~\cite{rong2020dropedge}, which randomly removes a portion of edges from the training adjacency matrix. Edge Proposal Sets (EPS)~\cite{singh2021edge}, which uses paired LP models to filter and rank top-k edges to insert as Common Neighbors in the training adjacency matrix. Lastly, we consider Topological Concentration (TC)~\cite{wang2023topological}, which re-weights the edges within the training adjacency matrix based on their shared information. \textit{Note:} We were unable to test CFLP~\cite{zhao2022learning} and EERM~\cite{wu2022eerm} after adapting their current implementations to the HeaRT evaluation setting. CFLP experienced an out-of-memory error on \textit{all} tested LPShift splits. EERM experienced an out-of-memory error on \textit{every} LPShift split of ogbl-ppa and ogbl-ddi, while also exceeding \textit{48 hours} per run on ogbl-collab without converging.

\begin{figure}[ht]
   \centering
   \includegraphics[width=\linewidth]{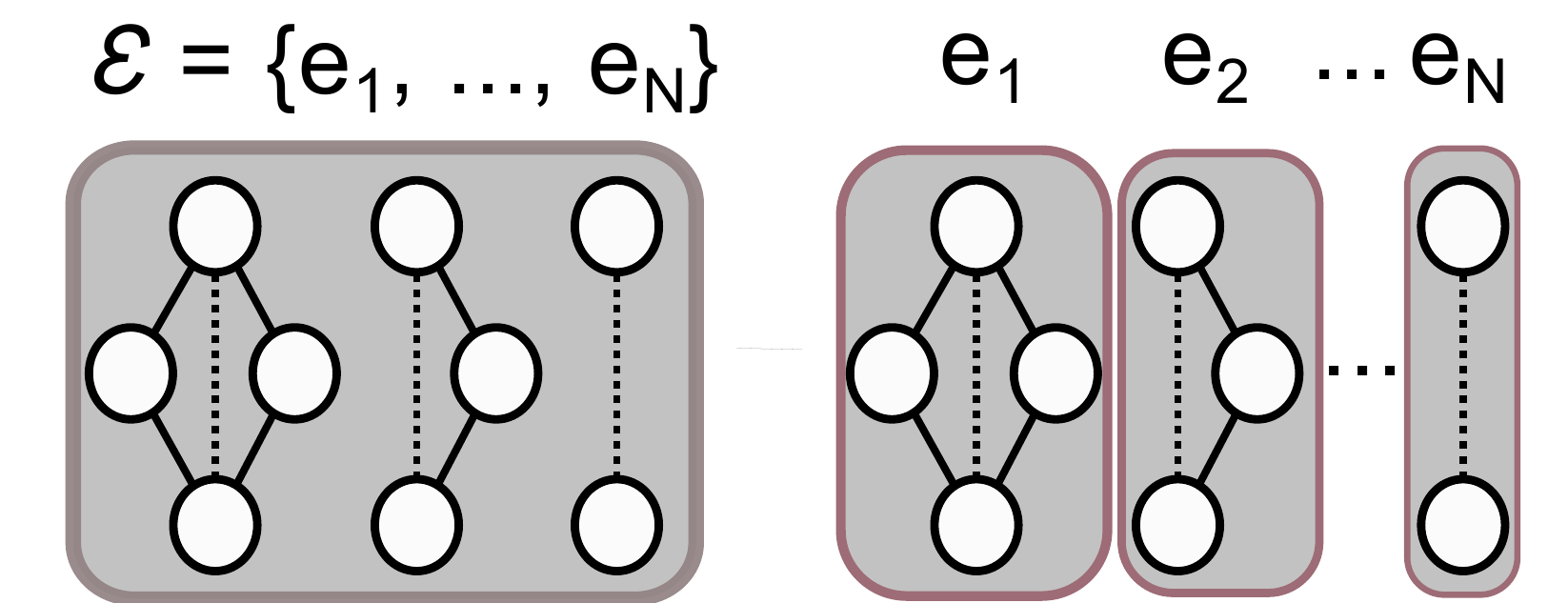}
   \caption{The original environment of dataset samples (shown left) split into subsets by their CN heuristic scores (shown right).}
   \Description{The original environment of dataset samples (shown left) split into subsets by their CN heuristic scores (shown right).}
   \label{fig:cn_subsets}
\end{figure}

{\bf Traditional Generalization Methods:} The final round of tests applies: IRM~\cite{arjovsky2019invariant}, VREx~\cite{krueger2021out}, GroupDRO~\cite{sagawa2019distributionally}, DANN~\cite{ganin2016domain}, 
 and DeepCORAL~\cite{sun2016deep} as generalization methods to GCN. Given that LPShift splits graph data directly on the sample structure, all tests with traditional generalization methods treat training samples as an environmental partition. The partitions \(\mathcal{E} = \{e_1,...,e_N\}\) are composed of $N$ environmental subsets, where each subset is binned based on their neighborhood heuristic; as demonstrated for CNs in Figure~\ref{fig:cn_subsets}. This binning of environmental subsets forces traditional generalization methods to target graph structure, controlling for the method's ability to improve GNN4LP performance under structural shift induced by LPShift.

{\bf Evaluation Setting}: We consider the standard evaluation procedure in LP, in which every positive validation/test sample is compared against $M$ negative samples. The goal is that the model should output a higher score (i.e., probability) for positive sample than the negatives. To create the negatives, we make use of the HeaRT evaluation setting \cite{li2024evaluating} which generates $M$ negatives samples {\it per positive sample} according to a set of common LP heuristics. In our study, we set $M=250$ and use CNs coupled with PPR as the heuristic in HeaRT.

{\bf Evaluation Metrics}: We evaluate all methods using the mean reciprocal rank (MRR) across multiple Hits@K ranking metrics, K=$\{1,3,5,10,20,50,100\}$.

\begin{table*}[ht]
\centering
\caption{The average percent change in MRR for each tuned model versus the HeaRT benchmark~\cite{li2024evaluating}, sorted by dataset then LPShift split and direction.}
\label{table:lpshift_vs_heart}
\begin{tabular}{l l l ccccccc}
\toprule
\textbf{Dataset} & \textbf{Split} & \textbf{Direction} & \textbf{GCN} & \textbf{RA} & \textbf{NCNC} & \textbf{NeoGNN} & \textbf{BUDDY} & \textbf{LPFormer} & \textbf{SEAL} \\
\midrule
\multirow{6}{*}{PPA} 
    & \multirow{2}{*}{CN} & Forward  & \colordown{-69.4\%}  & \colordown{-84.0\%}  & \colordown{-84.5\%}  & \colordown{-77.0\%}  & \colordown{-84.8\%}  & \colordown{-91.2\%}  & \colordown{-67.5\%} \\
    &                    & Backward & \colordown{-91.4\%}  & \colordown{-96.9\%}  & \colordown{-75.0\%}  & \colordown{-96.3\%}  & \colordown{-92.0\%}  & \colordown{-88.3\%}  & \colordown{-96.3\%} \\
\cmidrule(lr){2-10}
    & \multirow{2}{*}{SP} & Forward  & \colordown{-78.9\%}  & \colordown{-7.5\%}   & \colordown{-82.2\%}  & \colordown{-72.4\%}  & \colordown{-83.3\%}  & \colordown{-81.7\%}  & \colordown{-75.1\%} \\
    &                    & Backward & \colordown{-75.4\%}  & \colordown{-97.9\%}  & \colordown{-89.8\%}  & \colordown{-86.5\%}  & \colordown{-76.6\%}  & \colordown{-91.4\%}  & \colordown{-96.6\%} \\
\cmidrule(lr){2-10}
    & \multirow{2}{*}{PA} & Forward  & \colordown{-84.3\%}  & \colordown{-88.5\%}  & \colordown{-85.4\%}  & \colordown{-35.0\%}  & \colordown{-84.6\%}  & \colordown{-77.3\%}  & \colordown{-87.4\%} \\
    &                    & Backward & \colordown{-89.6\%}  & \colordown{-78.5\%}  & \colordown{-74.0\%}  & \colordown{-77.8\%}  & \colordown{-90.3\%}  & \colordown{-75.9\%}  & \colordown{-86.8\%} \\
\midrule
\multirow{4}{*}{DDI} 
    & \multirow{2}{*}{CN} & Forward  & \colordown{-66.1\%}  & \colordown{-50.5\%}  & >24hr   & \colordown{-85.3\%}  & \colordown{-82.4\%}  & \colordown{-93.1\%}  & \colordown{-89.9\%} \\
    &                    & Backward & \colordown{-91.2\%}  & \colordown{-93.9\%}  & >24hr   & \colordown{-80.7\%}  & \colordown{-91.2\%}  & \colordown{-73.4\%}  & \colordown{-93.0\%} \\
\cmidrule(lr){2-10}
    & \multirow{2}{*}{PA} & Forward  & \colordown{-84.0\%}  & \colordown{-56.2\%}  & >24hr   & \colordown{-86.0\%}  & \colordown{-85.3\%}  & \colordown{-97.1\%}  & \colordown{-87.8\%} \\
    &                    & Backward & \colordown{-80.5\%}  & \colordown{-60.8\%}  & >24hr   & \colordown{-75.1\%}  & \colordown{-83.9\%}  & \colordown{-74.4\%}  & \colordown{-74.4\%} \\
\midrule
\multirow{6}{*}{Collab} 
    & \multirow{2}{*}{CN} & Forward  & \colordown{-5.7\%}   & \colorup{+8.8\%}   & \colordown{-41.5\%}  & \colordown{-49.8\%}  & \colordown{-49.1\%}  & \colorup{+89.6\%}  & \colordown{-29.8\%} \\
    &                    & Backward & \colordown{-69.6\%}  & \colordown{-74.8\%}  & \colordown{-64.1\%}  & \colordown{-79.6\%}  & \colordown{-87.0\%}  & \colordown{-37.7\%}  & \colordown{-81.6\%} \\
\cmidrule(lr){2-10}
    & \multirow{2}{*}{SP} & Forward  & \colordown{-23.0\%}  & \colorup{+20.4\%}  & \colordown{-49.0\%}  & \colordown{-72.7\%}  & \colordown{-49.7\%}  & \colorup{+33.2\%}  & \colordown{-90.7\%} \\
    &                    & Backward & \colordown{-7.7\%}   & \colordown{-97.7\%}  & \colordown{-95.6\%}  & \colordown{-83.8\%}   & \colordown{-84.0\%}  & \colordown{-67.1\%}   & \colordown{-96.2\%} \\
\cmidrule(lr){2-10}
    & \multirow{2}{*}{PA} & Forward  & \colordown{-17.5\%}  & \colorup{+3.9\%}   & \colordown{-27.1\%}  & \colordown{-35.7\%}  & \colordown{-50.3\%}  & \colorup{+117.5\%}  & \colordown{-2.4\%} \\
    &                    & Backward & \colordown{-6.5\%}   & \colorup{+41.6\%}  & \colorup{+1.9\%}   & \colordown{-77.1\%}  & \colordown{-44.7\%}  & \colorup{+203.4\%} & \colorup{+52.9\%} \\
\bottomrule
\end{tabular}
\end{table*}

{\bf Hyperparameters}: All methods were tuned on permutations of learning rates in $\{1e^{-2}, 1e^{-3}\}$ and dropout in $\{0.1, 0.3\}$. Each model was trained and tested over five seeds to obtain the mean and standard deviations of their results. Given the significant time complexity of training and testing on LPShift variants for the ogbl-ppa datasets, NCNC and LPFormer were tuned on a single seed, followed by evaluation of the tuned model on five separate seeds. Additional hyperparameter tuning details are included in Appendix~\ref{sec:app_train_details}.

\subsection{Results for GNN4LP}\label{sec:gnn4lp_res}

In order to provide a unified perspective on how distribution shift affects link prediction models, GNN4LP models were trained and tested across five seeded runs on versions of ogbl-collab, ppa, and ddi split by: Common Neighbors, Shortest-Path, and Preferential-Attachment. Examining the results, we have the following three key observations.

{\bf \underline{Observation 1}: Drastic Changes in Model Performance}. As shown in Table~\ref{table:lpshift_vs_heart}, the majority of synthetic splits generated by LPShift reduce model performance significantly below the original HeaRT standard~\cite{li2024evaluating}. This is especially pronounced in denser graphs like ogbl-ppa and ogbl-ddi, indicating that LPShift affects the varied structures of graph data, which then reduces downstream perfomance. The only exception is ogbl-collab, where: RA, NCNC, LPFormer, and SEAL experience measurable improvements over the HeaRT baseline. Given ogbl-collab's lack of density relative to ogbl-ppa and ogbl-ddi, there are considerations about how LPShift induces a covariate shift to features~\cite{koh2021wilds}, especially where structure is correlated to the feature distribution. 

Additional analysis shown in Figure~\ref{fig:lpshift_collab_perf} indicates how dataset structure relates to the performance effects of LPShift.

\begin{figure}[h]
    \centering
    \includegraphics[width=\linewidth]{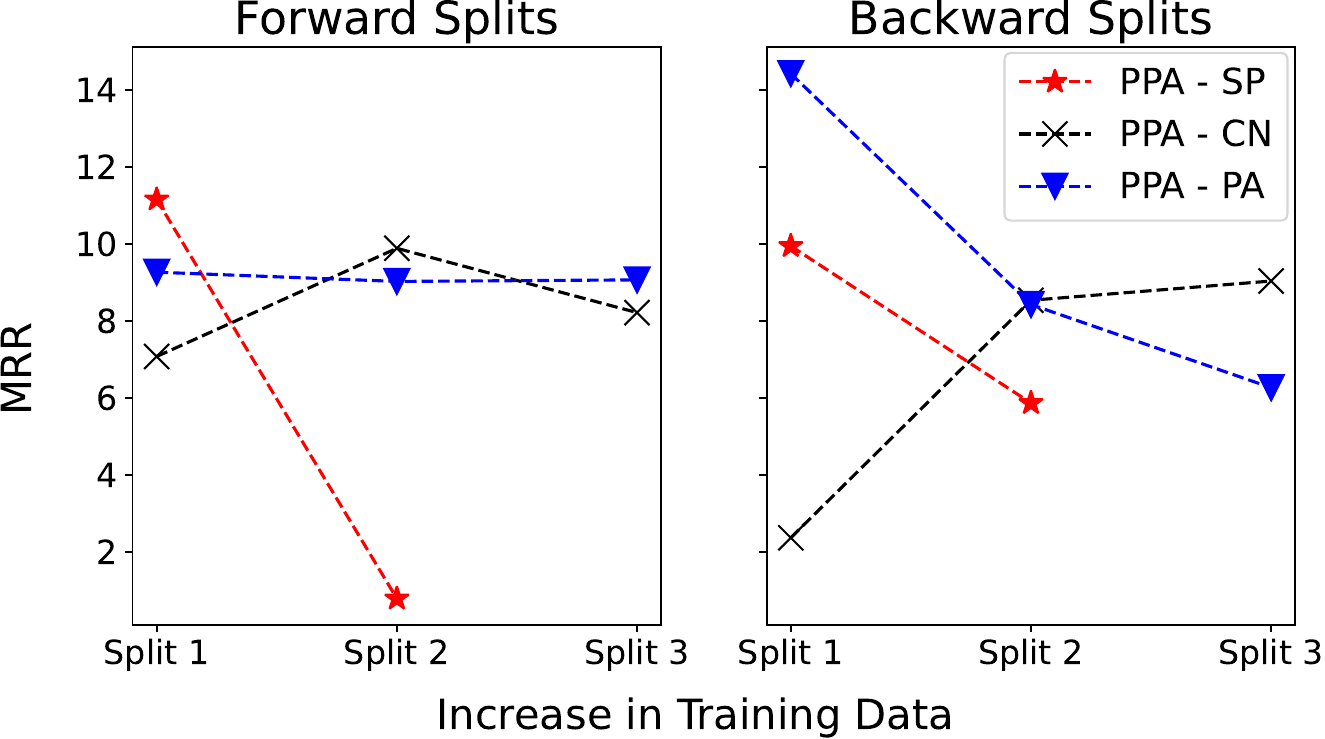}
    \caption{The mean scores of the best-performing GNN4LP models on LPShift's ogbl-ppa dataset. Each line represents a given dataset and split, arranged uniformly between figures. In the case of decreasing performance, the model with the highest average values was selected. }
    \Description{The mean scores of the best-performing GNN4LP models tested with our proposed splitting strategy. Each line represents a given dataset and split, arranged uniformly between figures. In the case of decreasing performance, the model with the highest average values was selected.}
    \label{fig:perform_splits}
\end{figure}

\textbf{\underline{Observation 2}: Performance Differs Across LPShift Splits.} As shown in Figure \ref{fig:perform_splits}, regardless of whether a model is tested on a ``Forward'' or ``Backward'' split of ogbl-ppa; the change in structural information for each subsequent split gradually changes a model's performance. On the ``CN'' split, a stark increase is seen between ''Split 1'' and ''Split 2'' (also detailed in descending order within Table~\ref{table:ppa_mrr}), indicating that more structural information between training and validation improves GNN4LP performance \cite{wang2023neural}, even on dense datasets like ogbl-ppa. The fact that these results include splits produced via Preferential-Attachment, Global Structural Information (SP), and Local Structural Information (CN) indicates the effect of \textit{any} change in structural information when training LP models \cite{mao2023revisiting}. In principle, Observation 1 and 2 answer RQ1 and RQ2 by indicating that structural alignment between positive samples from separate splits is critical for GNN4LP performance. LPShift makes this effect quantifiable through user-controlled thresholds, enabling future research to: induce, diagnose, and correct problems caused by LP-specific distribution shift.

\textbf{\underline{Observation 3}: Performance Impact Varies by Model.} All raw MRR scores are stored in Appendix~\ref{sec:app_dataset_res_tables} within ~\Crefrange{table:collab_mrr}{table:ddi_mrr}. From those tables, we can make the following observations by split type. \textbf{Common Neighbors:} Most models fail to generalize on the ``Backward'' CN splits. However, once more Common Neighbors are made available in the CN - 4,2 and CN - 5,3 splits; NCNC performs 2 to 3 times better than other GNN4LP models. Therefore, indicating that it is possible to generalize with limited local information. \textbf{Shortest-Path:} GNN4LP Models which rely more on local structural information (i.e. NCNC, LPFormer, and SEAL) typically suffer more under the ``Backward'' SP splits, resulting in the models performing 2x to 4x worse than BUDDY or GCN. Therefore, indicating the necessity for models to adapt in scenarios with an absence of local structural information. \textbf{Preferential-Attachment:} LPFormer performance on the PA split is 2 times higher than HeaRT's ogbl-collab~\cite{li2024evaluating}, but reduces drastically on LPShift's ogbl-ppa or ogbl-ddi. Therefore, indicating the impact that structural shift incurs with the increasing density of datasets. These observations within Section~\ref{sec:gnn4lp_res} provide implications for RQ2, showing that targeted model updates; like NCNC applying completion with CNs~\cite{wang2023neural} or LPFormer's PPR-aware attention~\citep{shomer2024lpformer, brin1998anatomy} aid generalization within CN and PA distribution shifts.


\subsection{Results for Generalization Methods}\label{sec:gen_meth_res}

In this section, we apply DropEdge \cite{rong2020dropedge}, EPS \cite{singh2021edge}, and TC \cite{zhao2023learning} on the previously benchmarked BUDDY \cite{chamberlain2022graph} to determine the feasibility of improving the LP models' generalization under LPShift.


\begin{table}[ht]
\centering
\caption{The raw MRR change for the BUDDY results in Table~\ref{table:lpshift_vs_heart}, after applying graph-specific generalization methods, averaged over LPShift splits.}
\begin{tabular}{lcccccc}
\toprule
 & \multicolumn{3}{c}{Forward} & \multicolumn{3}{c}{Backward} \\
 & CN & SP & PA & CN & SP & PA \\
\midrule
\multicolumn{7}{c}{Collab} \\
\midrule
ED & \colordown{-0.5} & \colordown{-1.6} & \colordown{-0.11} & \colordown{-0.72} & \colordown{-1.09} & \colorup{+0.36} \\
TC       & \colordown{-3.63} & \colordown{-8.02} & \colordown{-2.26} & \colorup{+1.29} & \colorup{+1.34} & \colordown{-7.38} \\
EPS      & \colordown{-4.28} & \colordown{-7.82} & \colordown{-4.46} & \colorup{+1.38} & \colorup{+1.45} & \colordown{-8.55} \\
\midrule
\multicolumn{7}{c}{PPA} \\
\midrule
ED & \colordown{-0.02} & \colorup{+1.31} & +0.003 & +0.003 & 0.0 & +0.006 \\
TC       & \colordown{-1.31} & \colorup{+0.86} & \colordown{-0.90} & \colorup{+1.27} & \colorup{+1.58} & \colordown{-1.09} \\
EPS      & \colordown{-1.12} & \colorup{+1.14} & \colordown{-1.02} & \colorup{+1.06} & \colorup{+2.2} & \colordown{-0.91} \\
\midrule
\multicolumn{7}{c}{DDI} \\
\midrule
ED & 0.0 & --    & 0.0 & 0.0 & --    & \colorup{+3.87} \\
TC       & \colordown{-1.17} & --    & 0.0 & \colordown{-0.12} & --    & \colordown{-0.46} \\
EPS      & \colordown{-0.86} & --    & \colordown{-0.58} & \colorup{+1.19} & --    & \colorup{+7.5} \\
\bottomrule
\end{tabular}

\label{table:gen_rel_change}
\end{table}

\begin{figure}[h]
    \centering
    \includegraphics[width=\linewidth]{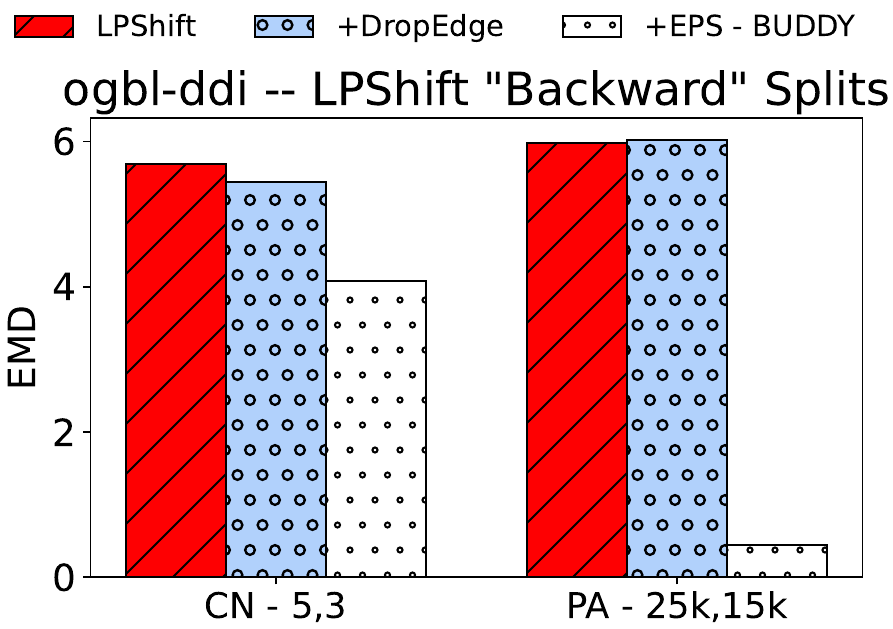}
    \caption{The EMD values calculated between the heuristic scores of training and testing samples. \textit{Note:} The tested heuristics correspond to their labelled splits, so as to simulate the dataset splitting. }
    \Description{The EMD values calculated between the heuristic scores of training and testing samples. \textit{Note:} The tested heuristics correspond to their labelled splits, so as to simulate the dataset splitting. }
    \label{fig:emd_ddi_plot_bw}
\end{figure}

{\bf \underline{Observation 1}: Graph-Specific Generalization Methods Can Help}. As demonstrated in Table~\ref{table:gen_rel_change}, the two generalization methods specific to LP: TC \cite{wang2023topological} and EPS \cite{singh2021edge} fail to increase performance under LPShift in the ''Forward'' CN and PA splits. However, the methods do improve performance for the "Backward" DDI split, where both EdgeDrop and EPS increase model performance by 3.87\% and 7.5\%, respectively. This is likely due to the exceptionally dense nature of DDI, where numerous high-degree nodes are likely made more relevant to performance through: 1) the removal of edges from EdgeDrop or 2) connecting relevant neighbors with EPS. To validate this, we calculate Earth Mover's Distance (EMD) \cite{rubner1998emd} between the heuristic scores of the training and testing splits before and after applying the generalization methods. EPS injects CNs into the training adjacency matrix, significantly altering the training and testing distributions. This drastic change is indicated in Figures~\ref{fig:emd_ddi_plot_bw} with the 28-92\% improvement of EMD scores between training and testing from applying EPS. Such a change in EMD, coupled with the Table~\ref{table:gen_rel_change} results, demonstrates that generalizing under LPShift can be achieved with targeted augmentations to the graph data structure. However, EdgeDrop has relatively-minimal effect on the EMD scores, indicating that non-targeted augmentations can generalize from highly-dense training subsets to less-dense test subsets.

TC often decreases performance on ''Forward'' splits. This is likely due to LPShift's distinct split thresholds; meaning there is limited structural overlap between sample distributions. As such, TC can't re-weight the training adjacency matrix for improved generalization to neighborhood information~\cite{wang2023topological,li2022ood}. This result runs contrary to current work, where re-weighting is effective for handling distribution shifts in other graph tasks~\cite{zhou2022model} and computer vision~\cite{fang2020rethinking}. However, the ''Backward'' CN and SP splits indicate consistent performance increases induced by TC.

{\bf \underline{Observation 2}: Traditional Generalization Methods Work in Limited Scenarios}. As shown in Table~\ref{table:ood_gcn}, any of the generalization methods tested on the ''Forward' split rarely improve performance. IRM and DRO often reduce performance the most, sometimes by more than 10\% MRR. All methods encounter more consistent performance improvements on the ''Backward'' ogbl-ppa and ogbl-ddi splits. Given that all generalization methods are trained on environmental subsets defined by structural heuristics, it follows that the traditional generalization methods improve performance under LPShift when given sample subsets with distinct and diverse structural patterns. Future work could explore this idea further, applying different heuristic scores to each environmental subset or extending invariant learning principles to work for the link prediction task~\cite{wu2024graph}. As shown in Table~\ref{table:ood_gcn}, IRM performs well on "Backward" ogbl-ddi splits but fails on "Backward" ogbl-collab splits. Indeed, there is no single generalization method which functions under all LPShift scenarios for each given dataset. This result is consistent with the application of OOD generalization techniques in other data domains~\cite{gulrajani2020search}. As such, this result also answers RQ3 by showing LPShift poses a challenge to current generalization methods.

\begin{table}[ht]
\centering
\caption{The raw MRR change for the GCN results in Table~\ref{table:lpshift_vs_heart}, after applying traditional generalization methods, averaged over LPShift splits.}
\begin{tabular}{lcccccc}
\toprule
 & \multicolumn{3}{c}{Forward} & \multicolumn{3}{c}{Backward} \\
 & CN & SP & PA & CN & SP & PA \\
\midrule
 \multicolumn{7}{c}{Collab}\\
 \midrule
IRM      & \colordown{-10.62} & \colordown{-8.51} & \colordown{-6.74} & \colordown{-2.66} & \colordown{-2.18} & \colordown{-14.86} \\
VREx     & \colordown{-0.53}  & \colorup{+0.39} & \colorup{+2.12} & \colorup{+0.31} & \colorup{+0.02} & \colordown{-1.29} \\
DRO & \colordown{-11.96} & \colordown{-7.76} & \colordown{-7.41} & \colordown{-2.51} & \colordown{-1.32} & \colordown{-13.66} \\
DANN     & \colordown{-0.13}  & \colorup{+0.74} & \colorup{+2.35} & \colorup{+0.39} & \colorup{+3.36} & \colordown{-1.75} \\
CORAL    & \colordown{-0.54}  & \colorup{+0.36} & \colorup{+2.02} & \colorup{+0.25} & \colorup{+3.32} & \colordown{-1.29} \\
\midrule
\multicolumn{7}{c}{PPA} \\
\midrule
IRM      & \colordown{-6.16} & \colorup{+0.02} & \colordown{-0.15}  & \colorup{+0.3} & \colordown{-4.51} & \colorup{+0.53} \\
VREx     & \colordown{-0.9} & \colorup{+0.08} & \colorup{+0.07} & \colorup{+0.79} & \colordown{-0.17}  & \colorup{+0.03} \\
DRO & \colordown{-5.86} & \colordown{-1.87} & \colordown{-0.73} & \colorup{+0.45} & \colordown{-4.48} & \colorup{+0.83} \\
DANN     & \colordown{-0.84} & \colorup{+0.13} & \colorup{+0.65} & \colorup{+0.72} & \colordown{-0.12}  & \colordown{-0.45} \\
CORAL    & \colordown{-0.91} & \colorup{+0.02} & \colorup{+0.37} & \colorup{+0.71} & \colordown{-0.17}  & \colordown{-0.51} \\
\midrule
\multicolumn{7}{c}{DDI} \\
\midrule
IRM      & \colordown{-3.44} & --      & \colordown{-0.56}  & \colorup{+3.24} & --      & \colorup{+3.26} \\
VREx     & \colordown{-3.68} & --     & \colordown{-0.66}  & \colorup{+2.76} & --      & \colorup{+1.56} \\
DRO & \colordown{-3.47} & --      & \colordown{-0.77}  & \colorup{+3.16} & --      & \colorup{+1.89} \\
DANN     & \colordown{-3.55} & --      & \colordown{-0.66}  & \colorup{+3.26} & --      & \colorup{+1.59} \\
CORAL    & \colordown{-3.72} & --      & \colordown{-0.62}  & \colorup{+2.74} & --      & \colorup{+1.18} \\
\bottomrule
\end{tabular}

\label{table:ood_gcn}
\end{table}

\subsection{Discussion}

{\bf Does GNN4LP generalize and do generalization methods work?} As detailed in Table~\ref{table:lpshift_vs_heart}, LPShift significantly reduces the performance for a majority of GNN4LP models versus the HeaRT baseline. This is especially notable given the difficulty HeaRT imposes on the original benchmark setting~\cite{hu2020open}. This observation, coupled with the limited success of all generalization methods on ''Forward'' splits in Tables~\ref{table:gen_rel_change} and~\ref{table:ood_gcn}, shows the difficulty posed on LP models when trained with limited structural information relative to evaluation samples.

\begin{figure}[h]
    \centering
    \includegraphics[width=\linewidth]{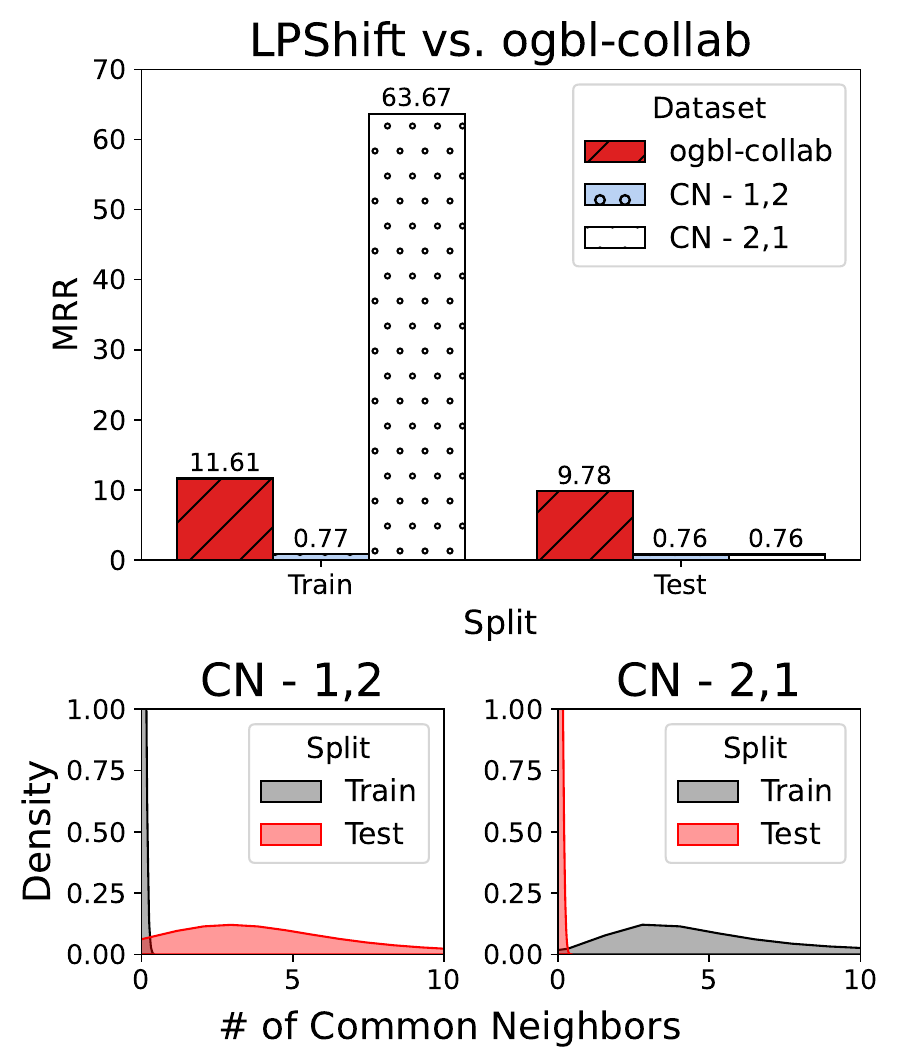}
    \caption{Three subplots corresponding to: Top.) CN predictor performance for the original ogbl-collab with it's respective LPShift 'CN - 1,2', and 'CN - 2,1' splits., Bottom-Left.) the density of Common Neighbors for 'CN - 1,2' LPShift on ogbl-collab, Bottom-Right.) the density of Common Neighbors for 'CN - 2,1' LPShift on ogbl-collab. }
    \Description{}
    \label{fig:lpshift_collab_perf}
\end{figure}

{\bf How is the proposed distribution shift affecting performance?}  In order to answer RQ4 and verify that LPShift's effect on dataset structure is significantly correlated to performance, Figure~\ref{fig:lpshift_collab_perf} follows a similar analysis conducted in \cite{wang2023neural}. In which, CN's predictive performance is measured and compared under the ''Forward'' and ''Backward'' LPShift. We choose CN as the predictor, given that directly ranks given dataset samples by structure to predict if a link forms; meaning it's predictive performance is sensitive to a change in dataset structure.
The bottom two subplots in Figure~\ref{fig:lpshift_collab_perf} depict the density estimates for CN distributions of LPShift's CN - 1,2 and CN - 2,1 splits on the ogbl-collab dataset. The topmost subplot depicts the performance of CN on the datasets depicted in the bottom two subplots, as well as the original ogbl-collab dataset.

Figure~\ref{fig:lpshift_collab_perf} demonstrates the targeted effect of LPShift on dataset structure, the reduced performance (1\% MRR) of the CN ranking on the CN - 1,2 split is due to the lack of structural information in the training split (CN = 0), limiting CN's ability to rank valid and test samples with $>$1 CNs. The structure-to-performance link is clearer in the CN - 2,1 scenario: where training with 2+ CNs yields 60\% MRR, but fails to generalize on test samples with zero CNs, shown by the 1\% MRR score. Given that all GNN4LP models incorporate neighborhood/structural information into their architectures~\citep{wang2023neural, chamberlain2022graph}, it follows that GNN4LP's reduced performance under LPShift is due to the splitting strategy enforcing a scenario where models must generalize to dataset samples drawn from a different distributions, empirically verifying Theorem~\ref{def:psi_theor}.

\section{Conclusion}

This work proposes LPShift, a simple dataset splitting strategy for inducing structural shift relevant for link prediction. The effect of this structural shift was then benchmarked on 16 shifted versions of ogbl-collab, ppa, and ddi, posing a unique challenge for SOTA GNN4LP models and generalization methods versus the HeaRT baseline. Further analysis indicates that current generalization methods improve performance under LPShift in scenarios with diverse pairwise structures amongst training samples. This indicates promising future directions for targeted graph data augmentation and invariant learning methods to improve generalization under distribution shift in the link prediction task. Regarding this, LPShift also demonstrates that GNN4LP as standalone models are vulnerable to distribution shift induced by pairwise heuristics. Given the severity of structural shift induced by LPShift, future work can consider relaxing the threshold between sample distributions or consider mixing structural heuristics to better mimic real-world distribution shift and the complexities therein. 


\begin{acks}
All authors are supported by the National Science Foundation (NSF) under grant numbers DGE2244164, CNS2321416, IIS2212032, IIS2212144, IOS2107215, DUE2234015, CNS2246050, DRL2405483 and IOS2035472, US Department of Commerce, Gates Foundation, the Michigan Department of Agriculture and Rural Development, Amazon, Meta, and SNAP.
\end{acks}

\bibliographystyle{ACM-Reference-Format}
\bibliography{sample-base}


\begin{thebibliography}{71}


\ifx \showCODEN    \undefined \def \showCODEN     #1{\unskip}     \fi
\ifx \showISBNx    \undefined \def \showISBNx     #1{\unskip}     \fi
\ifx \showISBNxiii \undefined \def \showISBNxiii  #1{\unskip}     \fi
\ifx \showISSN     \undefined \def \showISSN      #1{\unskip}     \fi
\ifx \showLCCN     \undefined \def \showLCCN      #1{\unskip}     \fi
\ifx \shownote     \undefined \def \shownote      #1{#1}          \fi
\ifx \showarticletitle \undefined \def \showarticletitle #1{#1}   \fi
\ifx \showURL      \undefined \def \showURL       {\relax}        \fi
\providecommand\bibfield[2]{#2}
\providecommand\bibinfo[2]{#2}
\providecommand\natexlab[1]{#1}
\providecommand\showeprint[2][]{arXiv:#2}

\bibitem[Abbas et~al\mbox{.}(2021)]%
        {abbas2021application}
\bibfield{author}{\bibinfo{person}{Khushnood Abbas}, \bibinfo{person}{Alireza Abbasi}, \bibinfo{person}{Shi Dong}, \bibinfo{person}{Ling Niu}, \bibinfo{person}{Laihang Yu}, \bibinfo{person}{Bolun Chen}, \bibinfo{person}{Shi-Min Cai}, {and} \bibinfo{person}{Qambar Hasan}.} \bibinfo{year}{2021}\natexlab{}.
\newblock \showarticletitle{Application of network link prediction in drug discovery}.
\newblock \bibinfo{journal}{\emph{BMC bioinformatics}}  \bibinfo{volume}{22} (\bibinfo{year}{2021}), \bibinfo{pages}{1--21}.
\newblock


\bibitem[Adamic and Adar(2003)]%
        {adamic2003friends}
\bibfield{author}{\bibinfo{person}{Lada~A Adamic} {and} \bibinfo{person}{Eytan Adar}.} \bibinfo{year}{2003}\natexlab{}.
\newblock \showarticletitle{Friends and neighbors on the web}.
\newblock \bibinfo{journal}{\emph{Social networks}} \bibinfo{volume}{25}, \bibinfo{number}{3} (\bibinfo{year}{2003}), \bibinfo{pages}{211--230}.
\newblock


\bibitem[Albert and Barabási(2002)]%
        {AlbertComNet2002}
\bibfield{author}{\bibinfo{person}{Réka Albert} {and} \bibinfo{person}{Albert-László Barabási}.} \bibinfo{year}{2002}\natexlab{}.
\newblock \showarticletitle{Statistical mechanics of complex networks}.
\newblock \bibinfo{journal}{\emph{Reviews of Modern Physics}} \bibinfo{volume}{74}, \bibinfo{number}{1} (\bibinfo{date}{Jan.} \bibinfo{year}{2002}), \bibinfo{pages}{47–97}.
\newblock
\showISSN{1539-0756}
\href{https://doi.org/10.1103/revmodphys.74.47}{doi:\nolinkurl{10.1103/revmodphys.74.47}}


\bibitem[Arjovsky et~al\mbox{.}(2019)]%
        {arjovsky2019invariant}
\bibfield{author}{\bibinfo{person}{Martin Arjovsky}, \bibinfo{person}{L{\'e}on Bottou}, \bibinfo{person}{Ishaan Gulrajani}, {and} \bibinfo{person}{David Lopez-Paz}.} \bibinfo{year}{2019}\natexlab{}.
\newblock \showarticletitle{Invariant risk minimization}.
\newblock \bibinfo{journal}{\emph{arXiv preprint arXiv:1907.02893}} (\bibinfo{year}{2019}).
\newblock


\bibitem[Barabási and Albert(1999)]%
        {BAGraphs}
\bibfield{author}{\bibinfo{person}{Albert-László Barabási} {and} \bibinfo{person}{Réka Albert}.} \bibinfo{year}{1999}\natexlab{}.
\newblock \showarticletitle{Emergence of Scaling in Random Networks}.
\newblock \bibinfo{journal}{\emph{Science}} \bibinfo{volume}{286}, \bibinfo{number}{5439} (\bibinfo{year}{1999}), \bibinfo{pages}{509--512}.
\newblock
\href{https://doi.org/10.1126/science.286.5439.509}{doi:\nolinkurl{10.1126/science.286.5439.509}}
\showeprint{https://www.science.org/doi/pdf/10.1126/science.286.5439.509}


\bibitem[Bevilacqua et~al\mbox{.}(2021)]%
        {bevilacqua2021size}
\bibfield{author}{\bibinfo{person}{Beatrice Bevilacqua}, \bibinfo{person}{Yangze Zhou}, {and} \bibinfo{person}{Bruno Ribeiro}.} \bibinfo{year}{2021}\natexlab{}.
\newblock \showarticletitle{Size-invariant graph representations for graph classification extrapolations}. In \bibinfo{booktitle}{\emph{International Conference on Machine Learning}}. PMLR, \bibinfo{pages}{837--851}.
\newblock


\bibitem[Brin and Page(1998)]%
        {brin1998anatomy}
\bibfield{author}{\bibinfo{person}{Sergey Brin} {and} \bibinfo{person}{Lawrence Page}.} \bibinfo{year}{1998}\natexlab{}.
\newblock \showarticletitle{The anatomy of a large-scale hypertextual web search engine}.
\newblock \bibinfo{journal}{\emph{Computer networks and ISDN systems}} \bibinfo{volume}{30}, \bibinfo{number}{1-7} (\bibinfo{year}{1998}), \bibinfo{pages}{107--117}.
\newblock


\bibitem[Bronstein et~al\mbox{.}(2021)]%
        {bronstein2021geometric}
\bibfield{author}{\bibinfo{person}{Michael~M Bronstein}, \bibinfo{person}{Joan Bruna}, \bibinfo{person}{Taco Cohen}, {and} \bibinfo{person}{Petar Veli{\v{c}}kovi{\'c}}.} \bibinfo{year}{2021}\natexlab{}.
\newblock \showarticletitle{Geometric deep learning: Grids, groups, graphs, geodesics, and gauges}.
\newblock \bibinfo{journal}{\emph{arXiv preprint arXiv:2104.13478}} (\bibinfo{year}{2021}).
\newblock


\bibitem[Chamberlain et~al\mbox{.}(2022)]%
        {chamberlain2022graph}
\bibfield{author}{\bibinfo{person}{Benjamin~Paul Chamberlain}, \bibinfo{person}{Sergey Shirobokov}, \bibinfo{person}{Emanuele Rossi}, \bibinfo{person}{Fabrizio Frasca}, \bibinfo{person}{Thomas Markovich}, \bibinfo{person}{Nils Hammerla}, \bibinfo{person}{Michael~M Bronstein}, {and} \bibinfo{person}{Max Hansmire}.} \bibinfo{year}{2022}\natexlab{}.
\newblock \showarticletitle{Graph Neural Networks for Link Prediction with Subgraph Sketching}.
\newblock \bibinfo{journal}{\emph{arXiv preprint arXiv:2209.15486}} (\bibinfo{year}{2022}).
\newblock


\bibitem[Ding et~al\mbox{.}(2021)]%
        {ding2021closer}
\bibfield{author}{\bibinfo{person}{Mucong Ding}, \bibinfo{person}{Kezhi Kong}, \bibinfo{person}{Jiuhai Chen}, \bibinfo{person}{John Kirchenbauer}, \bibinfo{person}{Micah Goldblum}, \bibinfo{person}{David Wipf}, \bibinfo{person}{Furong Huang}, {and} \bibinfo{person}{Tom Goldstein}.} \bibinfo{year}{2021}\natexlab{}.
\newblock \showarticletitle{A Closer Look at Distribution Shifts and Out-of-Distribution Generalization on Graphs}. In \bibinfo{booktitle}{\emph{NeurIPS 2021 Workshop on Distribution Shifts: Connecting Methods and Applications}}.
\newblock
\urldef\tempurl%
\url{https://openreview.net/forum?id=XvgPGWazqRH}
\showURL{%
\tempurl}


\bibitem[Dong et~al\mbox{.}(2022)]%
        {dong2022fakeedge}
\bibfield{author}{\bibinfo{person}{Kaiwen Dong}, \bibinfo{person}{Yijun Tian}, \bibinfo{person}{Zhichun Guo}, \bibinfo{person}{Yang Yang}, {and} \bibinfo{person}{Nitesh Chawla}.} \bibinfo{year}{2022}\natexlab{}.
\newblock \showarticletitle{Fakeedge: Alleviate dataset shift in link prediction}. In \bibinfo{booktitle}{\emph{Learning on Graphs Conference}}. PMLR, \bibinfo{pages}{56--1}.
\newblock


\bibitem[Evtushenko and Kleinberg(2021)]%
        {evtushenko2021paradox}
\bibfield{author}{\bibinfo{person}{Anna Evtushenko} {and} \bibinfo{person}{Jon Kleinberg}.} \bibinfo{year}{2021}\natexlab{}.
\newblock \showarticletitle{The paradox of second-order homophily in networks}.
\newblock \bibinfo{journal}{\emph{Scientific Reports}} \bibinfo{volume}{11}, \bibinfo{number}{1} (\bibinfo{year}{2021}), \bibinfo{pages}{13360}.
\newblock


\bibitem[Fan et~al\mbox{.}(2019)]%
        {fan2019graph}
\bibfield{author}{\bibinfo{person}{Wenqi Fan}, \bibinfo{person}{Yao Ma}, \bibinfo{person}{Qing Li}, \bibinfo{person}{Yuan He}, \bibinfo{person}{Eric Zhao}, \bibinfo{person}{Jiliang Tang}, {and} \bibinfo{person}{Dawei Yin}.} \bibinfo{year}{2019}\natexlab{}.
\newblock \showarticletitle{Graph neural networks for social recommendation}. In \bibinfo{booktitle}{\emph{The world wide web conference}}. \bibinfo{pages}{417--426}.
\newblock


\bibitem[Fang et~al\mbox{.}(2020)]%
        {fang2020rethinking}
\bibfield{author}{\bibinfo{person}{Tongtong Fang}, \bibinfo{person}{Nan Lu}, \bibinfo{person}{Gang Niu}, {and} \bibinfo{person}{Masashi Sugiyama}.} \bibinfo{year}{2020}\natexlab{}.
\newblock \showarticletitle{Rethinking importance weighting for deep learning under distribution shift}.
\newblock \bibinfo{journal}{\emph{Advances in neural information processing systems}}  \bibinfo{volume}{33} (\bibinfo{year}{2020}), \bibinfo{pages}{11996--12007}.
\newblock


\bibitem[Ganin et~al\mbox{.}(2016)]%
        {ganin2016domain}
\bibfield{author}{\bibinfo{person}{Yaroslav Ganin}, \bibinfo{person}{Evgeniya Ustinova}, \bibinfo{person}{Hana Ajakan}, \bibinfo{person}{Pascal Germain}, \bibinfo{person}{Hugo Larochelle}, \bibinfo{person}{Fran{\c{c}}ois Laviolette}, \bibinfo{person}{Mario March}, {and} \bibinfo{person}{Victor Lempitsky}.} \bibinfo{year}{2016}\natexlab{}.
\newblock \showarticletitle{Domain-adversarial training of neural networks}.
\newblock \bibinfo{journal}{\emph{Journal of machine learning research}} \bibinfo{volume}{17}, \bibinfo{number}{59} (\bibinfo{year}{2016}), \bibinfo{pages}{1--35}.
\newblock


\bibitem[Gao et~al\mbox{.}(2023)]%
        {gao2023alleviating}
\bibfield{author}{\bibinfo{person}{Yuan Gao}, \bibinfo{person}{Xiang Wang}, \bibinfo{person}{Xiangnan He}, \bibinfo{person}{Zhenguang Liu}, \bibinfo{person}{Huamin Feng}, {and} \bibinfo{person}{Yongdong Zhang}.} \bibinfo{year}{2023}\natexlab{}.
\newblock \showarticletitle{Alleviating structural distribution shift in graph anomaly detection}. In \bibinfo{booktitle}{\emph{Proceedings of the Sixteenth ACM International Conference on Web Search and Data Mining}}. \bibinfo{pages}{357--365}.
\newblock


\bibitem[Granovetter(1978)]%
        {granovetter1978threshold}
\bibfield{author}{\bibinfo{person}{Mark Granovetter}.} \bibinfo{year}{1978}\natexlab{}.
\newblock \showarticletitle{Threshold models of collective behavior}.
\newblock \bibinfo{journal}{\emph{American journal of sociology}} \bibinfo{volume}{83}, \bibinfo{number}{6} (\bibinfo{year}{1978}), \bibinfo{pages}{1420--1443}.
\newblock


\bibitem[Gui et~al\mbox{.}(2022)]%
        {gui2022good}
\bibfield{author}{\bibinfo{person}{Shurui Gui}, \bibinfo{person}{Xiner Li}, \bibinfo{person}{Limei Wang}, {and} \bibinfo{person}{Shuiwang Ji}.} \bibinfo{year}{2022}\natexlab{}.
\newblock \showarticletitle{Good: A graph out-of-distribution benchmark}.
\newblock \bibinfo{journal}{\emph{Advances in Neural Information Processing Systems}}  \bibinfo{volume}{35} (\bibinfo{year}{2022}), \bibinfo{pages}{2059--2073}.
\newblock


\bibitem[Gulrajani and Lopez-Paz(2020)]%
        {gulrajani2020search}
\bibfield{author}{\bibinfo{person}{Ishaan Gulrajani} {and} \bibinfo{person}{David Lopez-Paz}.} \bibinfo{year}{2020}\natexlab{}.
\newblock \showarticletitle{In search of lost domain generalization}.
\newblock \bibinfo{journal}{\emph{arXiv preprint arXiv:2007.01434}} (\bibinfo{year}{2020}).
\newblock


\bibitem[Guo et~al\mbox{.}(2023)]%
        {guo2023data}
\bibfield{author}{\bibinfo{person}{Yuxin Guo}, \bibinfo{person}{Cheng Yang}, \bibinfo{person}{Yuluo Chen}, \bibinfo{person}{Jixi Liu}, \bibinfo{person}{Chuan Shi}, {and} \bibinfo{person}{Junping Du}.} \bibinfo{year}{2023}\natexlab{}.
\newblock \showarticletitle{A Data-centric Framework to Endow Graph Neural Networks with Out-Of-Distribution Detection Ability}. In \bibinfo{booktitle}{\emph{Proceedings of the 29th ACM SIGKDD Conference on Knowledge Discovery and Data Mining}}. \bibinfo{pages}{638--648}.
\newblock


\bibitem[Hodges~Jr(1958)]%
        {hodges1958significance}
\bibfield{author}{\bibinfo{person}{JL Hodges~Jr}.} \bibinfo{year}{1958}\natexlab{}.
\newblock \showarticletitle{The significance probability of the Smirnov two-sample test}.
\newblock \bibinfo{journal}{\emph{Arkiv f{\"o}r matematik}} \bibinfo{volume}{3}, \bibinfo{number}{5} (\bibinfo{year}{1958}), \bibinfo{pages}{469--486}.
\newblock


\bibitem[Holme and Kim(2002)]%
        {holme2002growing}
\bibfield{author}{\bibinfo{person}{Petter Holme} {and} \bibinfo{person}{Beom~Jun Kim}.} \bibinfo{year}{2002}\natexlab{}.
\newblock \showarticletitle{Growing scale-free networks with tunable clustering}.
\newblock \bibinfo{journal}{\emph{Physical review E}} \bibinfo{volume}{65}, \bibinfo{number}{2} (\bibinfo{year}{2002}), \bibinfo{pages}{026107}.
\newblock


\bibitem[Hu et~al\mbox{.}(2020)]%
        {hu2020open}
\bibfield{author}{\bibinfo{person}{Weihua Hu}, \bibinfo{person}{Matthias Fey}, \bibinfo{person}{Marinka Zitnik}, \bibinfo{person}{Yuxiao Dong}, \bibinfo{person}{Hongyu Ren}, \bibinfo{person}{Bowen Liu}, \bibinfo{person}{Michele Catasta}, {and} \bibinfo{person}{Jure Leskovec}.} \bibinfo{year}{2020}\natexlab{}.
\newblock \showarticletitle{Open graph benchmark: Datasets for machine learning on graphs}.
\newblock \bibinfo{journal}{\emph{Advances in neural information processing systems}}  \bibinfo{volume}{33} (\bibinfo{year}{2020}), \bibinfo{pages}{22118--22133}.
\newblock


\bibitem[Ji et~al\mbox{.}(2022)]%
        {ji2022drugood}
\bibfield{author}{\bibinfo{person}{Yuanfeng Ji}, \bibinfo{person}{Lu Zhang}, \bibinfo{person}{Jiaxiang Wu}, \bibinfo{person}{Bingzhe Wu}, \bibinfo{person}{Long-Kai Huang}, \bibinfo{person}{Tingyang Xu}, \bibinfo{person}{Yu Rong}, \bibinfo{person}{Lanqing Li}, \bibinfo{person}{Jie Ren}, \bibinfo{person}{Ding Xue}, {et~al\mbox{.}}} \bibinfo{year}{2022}\natexlab{}.
\newblock \showarticletitle{DrugOOD: Out-of-Distribution (OOD) Dataset Curator and Benchmark for AI-aided Drug Discovery--A Focus on Affinity Prediction Problems with Noise Annotations}.
\newblock \bibinfo{journal}{\emph{arXiv preprint arXiv:2201.09637}} (\bibinfo{year}{2022}).
\newblock


\bibitem[Jin et~al\mbox{.}(2022)]%
        {jin2022empowering}
\bibfield{author}{\bibinfo{person}{Wei Jin}, \bibinfo{person}{Tong Zhao}, \bibinfo{person}{Jiayuan Ding}, \bibinfo{person}{Yozen Liu}, \bibinfo{person}{Jiliang Tang}, {and} \bibinfo{person}{Neil Shah}.} \bibinfo{year}{2022}\natexlab{}.
\newblock \showarticletitle{Empowering graph representation learning with test-time graph transformation}.
\newblock \bibinfo{journal}{\emph{arXiv preprint arXiv:2210.03561}} (\bibinfo{year}{2022}).
\newblock


\bibitem[Kipf and Welling(2016)]%
        {kipf2016variational}
\bibfield{author}{\bibinfo{person}{Thomas~N Kipf} {and} \bibinfo{person}{Max Welling}.} \bibinfo{year}{2016}\natexlab{}.
\newblock \showarticletitle{Variational graph auto-encoders}.
\newblock \bibinfo{journal}{\emph{arXiv preprint arXiv:1611.07308}} (\bibinfo{year}{2016}).
\newblock


\bibitem[Kipf and Welling(2017)]%
        {kipf2017semi}
\bibfield{author}{\bibinfo{person}{Thomas~N. Kipf} {and} \bibinfo{person}{Max Welling}.} \bibinfo{year}{2017}\natexlab{}.
\newblock \showarticletitle{Semi-Supervised Classification with Graph Convolutional Networks}. In \bibinfo{booktitle}{\emph{International Conference on Learning Representations (ICLR)}}.
\newblock


\bibitem[Koh et~al\mbox{.}(2021)]%
        {koh2021wilds}
\bibfield{author}{\bibinfo{person}{Pang~Wei Koh}, \bibinfo{person}{Shiori Sagawa}, \bibinfo{person}{Henrik Marklund}, \bibinfo{person}{Sang~Michael Xie}, \bibinfo{person}{Marvin Zhang}, \bibinfo{person}{Akshay Balsubramani}, \bibinfo{person}{Weihua Hu}, \bibinfo{person}{Michihiro Yasunaga}, \bibinfo{person}{Richard~Lanas Phillips}, \bibinfo{person}{Irena Gao}, {et~al\mbox{.}}} \bibinfo{year}{2021}\natexlab{}.
\newblock \showarticletitle{Wilds: A benchmark of in-the-wild distribution shifts}. In \bibinfo{booktitle}{\emph{International conference on machine learning}}. PMLR, \bibinfo{pages}{5637--5664}.
\newblock


\bibitem[Kov{\'a}cs et~al\mbox{.}(2019)]%
        {kovacs2019network}
\bibfield{author}{\bibinfo{person}{Istv{\'a}n~A Kov{\'a}cs}, \bibinfo{person}{Katja Luck}, \bibinfo{person}{Kerstin Spirohn}, \bibinfo{person}{Yang Wang}, \bibinfo{person}{Carl Pollis}, \bibinfo{person}{Sadie Schlabach}, \bibinfo{person}{Wenting Bian}, \bibinfo{person}{Dae-Kyum Kim}, \bibinfo{person}{Nishka Kishore}, \bibinfo{person}{Tong Hao}, {et~al\mbox{.}}} \bibinfo{year}{2019}\natexlab{}.
\newblock \showarticletitle{Network-based prediction of protein interactions}.
\newblock \bibinfo{journal}{\emph{Nature communications}} \bibinfo{volume}{10}, \bibinfo{number}{1} (\bibinfo{year}{2019}), \bibinfo{pages}{1240}.
\newblock


\bibitem[Krueger et~al\mbox{.}(2021)]%
        {krueger2021out}
\bibfield{author}{\bibinfo{person}{David Krueger}, \bibinfo{person}{Ethan Caballero}, \bibinfo{person}{Joern-Henrik Jacobsen}, \bibinfo{person}{Amy Zhang}, \bibinfo{person}{Jonathan Binas}, \bibinfo{person}{Dinghuai Zhang}, \bibinfo{person}{Remi Le~Priol}, {and} \bibinfo{person}{Aaron Courville}.} \bibinfo{year}{2021}\natexlab{}.
\newblock \showarticletitle{Out-of-distribution generalization via risk extrapolation (rex)}. In \bibinfo{booktitle}{\emph{International conference on machine learning}}. PMLR, \bibinfo{pages}{5815--5826}.
\newblock


\bibitem[Li et~al\mbox{.}(2022a)]%
        {li2022ood}
\bibfield{author}{\bibinfo{person}{Haoyang Li}, \bibinfo{person}{Xin Wang}, \bibinfo{person}{Ziwei Zhang}, {and} \bibinfo{person}{Wenwu Zhu}.} \bibinfo{year}{2022}\natexlab{a}.
\newblock \showarticletitle{Ood-gnn: Out-of-distribution generalized graph neural network}.
\newblock \bibinfo{journal}{\emph{IEEE Transactions on Knowledge and Data Engineering}} (\bibinfo{year}{2022}).
\newblock


\bibitem[Li et~al\mbox{.}(2022b)]%
        {li2022out}
\bibfield{author}{\bibinfo{person}{Haoyang Li}, \bibinfo{person}{Xin Wang}, \bibinfo{person}{Ziwei Zhang}, {and} \bibinfo{person}{Wenwu Zhu}.} \bibinfo{year}{2022}\natexlab{b}.
\newblock \showarticletitle{Out-of-distribution generalization on graphs: A survey}.
\newblock \bibinfo{journal}{\emph{arXiv preprint arXiv:2202.07987}} (\bibinfo{year}{2022}).
\newblock


\bibitem[Li et~al\mbox{.}(2024)]%
        {li2024evaluating}
\bibfield{author}{\bibinfo{person}{Juanhui Li}, \bibinfo{person}{Harry Shomer}, \bibinfo{person}{Haitao Mao}, \bibinfo{person}{Shenglai Zeng}, \bibinfo{person}{Yao Ma}, \bibinfo{person}{Neil Shah}, \bibinfo{person}{Jiliang Tang}, {and} \bibinfo{person}{Dawei Yin}.} \bibinfo{year}{2024}\natexlab{}.
\newblock \showarticletitle{Evaluating graph neural networks for link prediction: Current pitfalls and new benchmarking}.
\newblock \bibinfo{journal}{\emph{Advances in Neural Information Processing Systems}}  \bibinfo{volume}{36} (\bibinfo{year}{2024}).
\newblock


\bibitem[Liben-Nowell and Kleinberg(2003)]%
        {liben2003link}
\bibfield{author}{\bibinfo{person}{David Liben-Nowell} {and} \bibinfo{person}{Jon Kleinberg}.} \bibinfo{year}{2003}\natexlab{}.
\newblock \showarticletitle{The link prediction problem for social networks}. In \bibinfo{booktitle}{\emph{Proceedings of the twelfth international conference on Information and knowledge management}}. \bibinfo{pages}{556--559}.
\newblock


\bibitem[Lin et~al\mbox{.}(2015)]%
        {lin2015learning}
\bibfield{author}{\bibinfo{person}{Yankai Lin}, \bibinfo{person}{Zhiyuan Liu}, \bibinfo{person}{Maosong Sun}, \bibinfo{person}{Yang Liu}, {and} \bibinfo{person}{Xuan Zhu}.} \bibinfo{year}{2015}\natexlab{}.
\newblock \showarticletitle{Learning entity and relation embeddings for knowledge graph completion}. In \bibinfo{booktitle}{\emph{Proceedings of the AAAI conference on artificial intelligence}}, Vol.~\bibinfo{volume}{29}.
\newblock


\bibitem[Mao et~al\mbox{.}(2024a)]%
        {mao2024demystifying}
\bibfield{author}{\bibinfo{person}{Haitao Mao}, \bibinfo{person}{Zhikai Chen}, \bibinfo{person}{Wei Jin}, \bibinfo{person}{Haoyu Han}, \bibinfo{person}{Yao Ma}, \bibinfo{person}{Tong Zhao}, \bibinfo{person}{Neil Shah}, {and} \bibinfo{person}{Jiliang Tang}.} \bibinfo{year}{2024}\natexlab{a}.
\newblock \showarticletitle{Demystifying Structural Disparity in Graph Neural Networks: Can One Size Fit All?}
\newblock \bibinfo{journal}{\emph{Advances in Neural Information Processing Systems}}  \bibinfo{volume}{36} (\bibinfo{year}{2024}).
\newblock


\bibitem[Mao et~al\mbox{.}(2024b)]%
        {mao2023revisiting}
\bibfield{author}{\bibinfo{person}{Haitao Mao}, \bibinfo{person}{Juanhui Li}, \bibinfo{person}{Harry Shomer}, \bibinfo{person}{Bingheng Li}, \bibinfo{person}{Wenqi Fan}, \bibinfo{person}{Yao Ma}, \bibinfo{person}{Tong Zhao}, \bibinfo{person}{Neil Shah}, {and} \bibinfo{person}{Jiliang Tang}.} \bibinfo{year}{2024}\natexlab{b}.
\newblock \showarticletitle{Revisiting Link Prediction: a data perspective}. In \bibinfo{booktitle}{\emph{The Twelfth International Conference on Learning Representations}}.
\newblock


\bibitem[Newman(2001)]%
        {newman2001clustering}
\bibfield{author}{\bibinfo{person}{Mark~EJ Newman}.} \bibinfo{year}{2001}\natexlab{}.
\newblock \showarticletitle{Clustering and preferential attachment in growing networks}.
\newblock \bibinfo{journal}{\emph{Physical review E}} \bibinfo{volume}{64}, \bibinfo{number}{2} (\bibinfo{year}{2001}), \bibinfo{pages}{025102}.
\newblock


\bibitem[Rong et~al\mbox{.}(2020)]%
        {rong2020dropedge}
\bibfield{author}{\bibinfo{person}{Yu Rong}, \bibinfo{person}{Wenbing Huang}, \bibinfo{person}{Tingyang Xu}, {and} \bibinfo{person}{Junzhou Huang}.} \bibinfo{year}{2020}\natexlab{}.
\newblock \showarticletitle{DropEdge: Towards Deep Graph Convolutional Networks on Node Classification}. In \bibinfo{booktitle}{\emph{International Conference on Learning Representations}}.
\newblock
\urldef\tempurl%
\url{https://openreview.net/forum?id=Hkx1qkrKPr}
\showURL{%
\tempurl}


\bibitem[Rubner et~al\mbox{.}(1998)]%
        {rubner1998emd}
\bibfield{author}{\bibinfo{person}{Y. Rubner}, \bibinfo{person}{C. Tomasi}, {and} \bibinfo{person}{L.J. Guibas}.} \bibinfo{year}{1998}\natexlab{}.
\newblock \showarticletitle{A metric for distributions with applications to image databases}. In \bibinfo{booktitle}{\emph{Sixth International Conference on Computer Vision (IEEE Cat. No.98CH36271)}}. \bibinfo{pages}{59--66}.
\newblock
\href{https://doi.org/10.1109/ICCV.1998.710701}{doi:\nolinkurl{10.1109/ICCV.1998.710701}}


\bibitem[Russell and Norvig(2009)]%
        {AIMABook}
\bibfield{author}{\bibinfo{person}{Stuart Russell} {and} \bibinfo{person}{Peter Norvig}.} \bibinfo{year}{2009}\natexlab{}.
\newblock \bibinfo{booktitle}{\emph{Artificial Intelligence: A Modern Approach} (\bibinfo{edition}{3rd} ed.)}.
\newblock \bibinfo{publisher}{Prentice Hall Press}, \bibinfo{address}{USA}.
\newblock
\showISBNx{0136042597}


\bibitem[Sagawa et~al\mbox{.}(2019)]%
        {sagawa2019distributionally}
\bibfield{author}{\bibinfo{person}{Shiori Sagawa}, \bibinfo{person}{Pang~Wei Koh}, \bibinfo{person}{Tatsunori~B Hashimoto}, {and} \bibinfo{person}{Percy Liang}.} \bibinfo{year}{2019}\natexlab{}.
\newblock \showarticletitle{Distributionally robust neural networks for group shifts: On the importance of regularization for worst-case generalization}.
\newblock \bibinfo{journal}{\emph{arXiv preprint arXiv:1911.08731}} (\bibinfo{year}{2019}).
\newblock


\bibitem[Schelling(1978)]%
        {schelling1978micromotives}
\bibfield{author}{\bibinfo{person}{Thomas~C Schelling}.} \bibinfo{year}{1978}\natexlab{}.
\newblock \showarticletitle{Micromotives and Macrobehavior WW Norton \& Company}.
\newblock \bibinfo{journal}{\emph{New York, NY}} (\bibinfo{year}{1978}).
\newblock


\bibitem[Shi et~al\mbox{.}(2023)]%
        {shi2023label}
\bibfield{author}{\bibinfo{person}{Zhihao Shi}, \bibinfo{person}{Jie Wang}, \bibinfo{person}{Fanghua Lu}, \bibinfo{person}{Hanzhu Chen}, \bibinfo{person}{Defu Lian}, \bibinfo{person}{Zheng Wang}, \bibinfo{person}{Jieping Ye}, {and} \bibinfo{person}{Feng Wu}.} \bibinfo{year}{2023}\natexlab{}.
\newblock \showarticletitle{Label Deconvolution for Node Representation Learning on Large-scale Attributed Graphs against Learning Bias}.
\newblock \bibinfo{journal}{\emph{arXiv preprint arXiv:2309.14907}} (\bibinfo{year}{2023}).
\newblock


\bibitem[Shomer et~al\mbox{.}(2024)]%
        {shomer2024lpformer}
\bibfield{author}{\bibinfo{person}{Harry Shomer}, \bibinfo{person}{Yao Ma}, \bibinfo{person}{Haitao Mao}, \bibinfo{person}{Juanhui Li}, \bibinfo{person}{Bo Wu}, {and} \bibinfo{person}{Jiliang Tang}.} \bibinfo{year}{2024}\natexlab{}.
\newblock \showarticletitle{LPFormer: an adaptive graph transformer for link prediction}. In \bibinfo{booktitle}{\emph{Proceedings of the 30th ACM SIGKDD Conference on Knowledge Discovery and Data Mining}}. \bibinfo{pages}{2686--2698}.
\newblock


\bibitem[Singh et~al\mbox{.}(2021)]%
        {singh2021edge}
\bibfield{author}{\bibinfo{person}{Abhay Singh}, \bibinfo{person}{Qian Huang}, \bibinfo{person}{Sijia~Linda Huang}, \bibinfo{person}{Omkar Bhalerao}, \bibinfo{person}{Horace He}, \bibinfo{person}{Ser-Nam Lim}, {and} \bibinfo{person}{Austin~R Benson}.} \bibinfo{year}{2021}\natexlab{}.
\newblock \showarticletitle{Edge proposal sets for link prediction}.
\newblock \bibinfo{journal}{\emph{arXiv preprint arXiv:2106.15810}} (\bibinfo{year}{2021}).
\newblock


\bibitem[Srinivasan and Ribeiro(2019)]%
        {srinivasan2019equivalence}
\bibfield{author}{\bibinfo{person}{Balasubramaniam Srinivasan} {and} \bibinfo{person}{Bruno Ribeiro}.} \bibinfo{year}{2019}\natexlab{}.
\newblock \showarticletitle{On the Equivalence between Positional Node Embeddings and Structural Graph Representations}. In \bibinfo{booktitle}{\emph{International Conference on Learning Representations}}.
\newblock


\bibitem[Sui et~al\mbox{.}(2024)]%
        {sui2024unleashing}
\bibfield{author}{\bibinfo{person}{Yongduo Sui}, \bibinfo{person}{Qitian Wu}, \bibinfo{person}{Jiancan Wu}, \bibinfo{person}{Qing Cui}, \bibinfo{person}{Longfei Li}, \bibinfo{person}{Jun Zhou}, \bibinfo{person}{Xiang Wang}, {and} \bibinfo{person}{Xiangnan He}.} \bibinfo{year}{2024}\natexlab{}.
\newblock \showarticletitle{Unleashing the power of graph data augmentation on covariate distribution shift}.
\newblock \bibinfo{journal}{\emph{Advances in Neural Information Processing Systems}}  \bibinfo{volume}{36} (\bibinfo{year}{2024}).
\newblock


\bibitem[Sun and Saenko(2016)]%
        {sun2016deep}
\bibfield{author}{\bibinfo{person}{Baochen Sun} {and} \bibinfo{person}{Kate Saenko}.} \bibinfo{year}{2016}\natexlab{}.
\newblock \showarticletitle{Deep coral: Correlation alignment for deep domain adaptation}. In \bibinfo{booktitle}{\emph{Computer Vision--ECCV 2016 Workshops: Amsterdam, The Netherlands, October 8-10 and 15-16, 2016, Proceedings, Part III 14}}. Springer, \bibinfo{pages}{443--450}.
\newblock


\bibitem[Wang et~al\mbox{.}(2023a)]%
        {wang2023neural}
\bibfield{author}{\bibinfo{person}{Xiyuan Wang}, \bibinfo{person}{Haotong Yang}, {and} \bibinfo{person}{Muhan Zhang}.} \bibinfo{year}{2023}\natexlab{a}.
\newblock \showarticletitle{Neural Common Neighbor with Completion for Link Prediction}. In \bibinfo{booktitle}{\emph{The Twelfth International Conference on Learning Representations}}.
\newblock


\bibitem[Wang et~al\mbox{.}(2023b)]%
        {wang2023topological}
\bibfield{author}{\bibinfo{person}{Yu Wang}, \bibinfo{person}{Tong Zhao}, \bibinfo{person}{Yuying Zhao}, \bibinfo{person}{Yunchao Liu}, \bibinfo{person}{Xueqi Cheng}, \bibinfo{person}{Neil Shah}, {and} \bibinfo{person}{Tyler Derr}.} \bibinfo{year}{2023}\natexlab{b}.
\newblock \showarticletitle{A Topological Perspective on Demystifying GNN-Based Link Prediction Performance}.
\newblock \bibinfo{journal}{\emph{arXiv preprint arXiv:2310.04612}} (\bibinfo{year}{2023}).
\newblock


\bibitem[Wei et~al\mbox{.}(2021)]%
        {PASGraph}
\bibfield{author}{\bibinfo{person}{Lanning Wei}, \bibinfo{person}{Huan Zhao}, \bibinfo{person}{Quanming Yao}, {and} \bibinfo{person}{Zhiqiang He}.} \bibinfo{year}{2021}\natexlab{}.
\newblock \showarticletitle{Pooling Architecture Search for Graph Classification}. In \bibinfo{booktitle}{\emph{Proceedings of the 30th ACM International Conference on Information and Knowledge Management}} \emph{(\bibinfo{series}{CIKM ’21})}. \bibinfo{publisher}{ACM}.
\newblock
\href{https://doi.org/10.1145/3459637.3482285}{doi:\nolinkurl{10.1145/3459637.3482285}}


\bibitem[Wiles et~al\mbox{.}(2021)]%
        {finedistshift}
\bibfield{author}{\bibinfo{person}{Olivia Wiles}, \bibinfo{person}{Sven Gowal}, \bibinfo{person}{Florian Stimberg}, \bibinfo{person}{Sylvestre-Alvise Rebuffi}, \bibinfo{person}{Ira Ktena}, \bibinfo{person}{Krishnamurthy Dvijotham}, {and} \bibinfo{person}{A. Cemgil}.} \bibinfo{year}{2021}\natexlab{}.
\newblock \showarticletitle{A Fine-Grained Analysis on Distribution Shift}.
\newblock \bibinfo{journal}{\emph{ArXiv}}  \bibinfo{volume}{abs/2110.11328} (\bibinfo{year}{2021}).
\newblock


\bibitem[Wu et~al\mbox{.}(2023)]%
        {wu2023energy}
\bibfield{author}{\bibinfo{person}{Qitian Wu}, \bibinfo{person}{Yiting Chen}, \bibinfo{person}{Chenxiao Yang}, {and} \bibinfo{person}{Junchi Yan}.} \bibinfo{year}{2023}\natexlab{}.
\newblock \showarticletitle{Energy-based out-of-distribution detection for graph neural networks}.
\newblock \bibinfo{journal}{\emph{arXiv preprint arXiv:2302.02914}} (\bibinfo{year}{2023}).
\newblock


\bibitem[Wu et~al\mbox{.}(2024)]%
        {wu2024graph}
\bibfield{author}{\bibinfo{person}{Qitian Wu}, \bibinfo{person}{Fan Nie}, \bibinfo{person}{Chenxiao Yang}, \bibinfo{person}{Tianyi Bao}, {and} \bibinfo{person}{Junchi Yan}.} \bibinfo{year}{2024}\natexlab{}.
\newblock \showarticletitle{Graph Out-of-Distribution Generalization via Causal Intervention}. In \bibinfo{booktitle}{\emph{Proceedings of the ACM on Web Conference 2024}}. \bibinfo{pages}{850--860}.
\newblock


\bibitem[Wu et~al\mbox{.}(2022)]%
        {wu2022eerm}
\bibfield{author}{\bibinfo{person}{Qitian Wu}, \bibinfo{person}{Hengrui Zhang}, \bibinfo{person}{Junchi Yan}, {and} \bibinfo{person}{David Wipf}.} \bibinfo{year}{2022}\natexlab{}.
\newblock \showarticletitle{Handling Distribution Shifts on Graphs: An Invariance Perspective}. In \bibinfo{booktitle}{\emph{International Conference on Learning Representations (ICLR)}}.
\newblock


\bibitem[Yao et~al\mbox{.}(2022a)]%
        {wildtime}
\bibfield{author}{\bibinfo{person}{Huaxiu Yao}, \bibinfo{person}{Caroline Choi}, \bibinfo{person}{Bochuan Cao}, \bibinfo{person}{Yoonho Lee}, \bibinfo{person}{Pang Wei~W Koh}, {and} \bibinfo{person}{Chelsea Finn}.} \bibinfo{year}{2022}\natexlab{a}.
\newblock \showarticletitle{Wild-Time: A Benchmark of in-the-Wild Distribution Shift over Time}. In \bibinfo{booktitle}{\emph{Advances in Neural Information Processing Systems}}, \bibfield{editor}{\bibinfo{person}{S.~Koyejo}, \bibinfo{person}{S.~Mohamed}, \bibinfo{person}{A.~Agarwal}, \bibinfo{person}{D.~Belgrave}, \bibinfo{person}{K.~Cho}, {and} \bibinfo{person}{A.~Oh}} (Eds.), Vol.~\bibinfo{volume}{35}. \bibinfo{publisher}{Curran Associates, Inc.}, \bibinfo{pages}{10309--10324}.
\newblock
\urldef\tempurl%
\url{https://proceedings.neurips.cc/paper_files/paper/2022/file/43119db5d59f07cc08fca7ba6820179a-Paper-Datasets_and_Benchmarks.pdf}
\showURL{%
\tempurl}


\bibitem[Yao et~al\mbox{.}(2022b)]%
        {yao2022sela}
\bibfield{author}{\bibinfo{person}{Huaxiu Yao}, \bibinfo{person}{Yu Wang}, \bibinfo{person}{Sai Li}, \bibinfo{person}{Linjun Zhang}, \bibinfo{person}{Weixin Liang}, \bibinfo{person}{James Zou}, {and} \bibinfo{person}{Chelsea Finn}.} \bibinfo{year}{2022}\natexlab{b}.
\newblock \showarticletitle{Improving Out-of-Distribution Robustness via Selective Augmentation}. In \bibinfo{booktitle}{\emph{Proceedings of the 39th International Conference on Machine Learning}} \emph{(\bibinfo{series}{Proceedings of Machine Learning Research}, Vol.~\bibinfo{volume}{162})}, \bibfield{editor}{\bibinfo{person}{Kamalika Chaudhuri}, \bibinfo{person}{Stefanie Jegelka}, \bibinfo{person}{Le~Song}, \bibinfo{person}{Csaba Szepesvari}, \bibinfo{person}{Gang Niu}, {and} \bibinfo{person}{Sivan Sabato}} (Eds.). \bibinfo{publisher}{PMLR}, \bibinfo{pages}{25407--25437}.
\newblock
\urldef\tempurl%
\url{https://proceedings.mlr.press/v162/yao22b.html}
\showURL{%
\tempurl}


\bibitem[Yuan et~al\mbox{.}(2021)]%
        {yuan2021largescale}
\bibfield{author}{\bibinfo{person}{Zhuoning Yuan}, \bibinfo{person}{Yan Yan}, \bibinfo{person}{Milan Sonka}, {and} \bibinfo{person}{Tianbao Yang}.} \bibinfo{year}{2021}\natexlab{}.
\newblock \bibinfo{title}{Large-scale Robust Deep AUC Maximization: A New Surrogate Loss and Empirical Studies on Medical Image Classification}.
\newblock
\showeprint[arxiv]{2012.03173}~[cs.LG]


\bibitem[Yun et~al\mbox{.}(2021)]%
        {yun2021neo}
\bibfield{author}{\bibinfo{person}{Seongjun Yun}, \bibinfo{person}{Seoyoon Kim}, \bibinfo{person}{Junhyun Lee}, \bibinfo{person}{Jaewoo Kang}, {and} \bibinfo{person}{Hyunwoo~J Kim}.} \bibinfo{year}{2021}\natexlab{}.
\newblock \showarticletitle{Neo-gnns: Neighborhood overlap-aware graph neural networks for link prediction}.
\newblock \bibinfo{journal}{\emph{Advances in Neural Information Processing Systems}}  \bibinfo{volume}{34} (\bibinfo{year}{2021}), \bibinfo{pages}{13683--13694}.
\newblock


\bibitem[Ze{\v{c}}evi{\'c} et~al\mbox{.}(2021)]%
        {zevcevic2021relating}
\bibfield{author}{\bibinfo{person}{Matej Ze{\v{c}}evi{\'c}}, \bibinfo{person}{Devendra~Singh Dhami}, \bibinfo{person}{Petar Veli{\v{c}}kovi{\'c}}, {and} \bibinfo{person}{Kristian Kersting}.} \bibinfo{year}{2021}\natexlab{}.
\newblock \showarticletitle{Relating graph neural networks to structural causal models}.
\newblock \bibinfo{journal}{\emph{arXiv preprint arXiv:2109.04173}} (\bibinfo{year}{2021}).
\newblock


\bibitem[Zhang and Chen(2018)]%
        {zhang2018link}
\bibfield{author}{\bibinfo{person}{Muhan Zhang} {and} \bibinfo{person}{Yixin Chen}.} \bibinfo{year}{2018}\natexlab{}.
\newblock \showarticletitle{Link prediction based on graph neural networks}.
\newblock \bibinfo{journal}{\emph{Advances in neural information processing systems}}  \bibinfo{volume}{31} (\bibinfo{year}{2018}).
\newblock


\bibitem[Zhang et~al\mbox{.}(2021)]%
        {zhang2021labeling}
\bibfield{author}{\bibinfo{person}{Muhan Zhang}, \bibinfo{person}{Pan Li}, \bibinfo{person}{Yinglong Xia}, \bibinfo{person}{Kai Wang}, {and} \bibinfo{person}{Long Jin}.} \bibinfo{year}{2021}\natexlab{}.
\newblock \showarticletitle{Labeling trick: A theory of using graph neural networks for multi-node representation learning}.
\newblock \bibinfo{journal}{\emph{Advances in Neural Information Processing Systems}}  \bibinfo{volume}{34} (\bibinfo{year}{2021}), \bibinfo{pages}{9061--9073}.
\newblock


\bibitem[Zhang et~al\mbox{.}(2022)]%
        {zhang2022dynamic}
\bibfield{author}{\bibinfo{person}{Zeyang Zhang}, \bibinfo{person}{Xin Wang}, \bibinfo{person}{Ziwei Zhang}, \bibinfo{person}{Haoyang Li}, \bibinfo{person}{Zhou Qin}, {and} \bibinfo{person}{Wenwu Zhu}.} \bibinfo{year}{2022}\natexlab{}.
\newblock \showarticletitle{Dynamic graph neural networks under spatio-temporal distribution shift}.
\newblock \bibinfo{journal}{\emph{Advances in neural information processing systems}}  \bibinfo{volume}{35} (\bibinfo{year}{2022}), \bibinfo{pages}{6074--6089}.
\newblock


\bibitem[Zhao et~al\mbox{.}(2022)]%
        {zhao2022learning}
\bibfield{author}{\bibinfo{person}{Jianan Zhao}, \bibinfo{person}{Meng Qu}, \bibinfo{person}{Chaozhuo Li}, \bibinfo{person}{Hao Yan}, \bibinfo{person}{Qian Liu}, \bibinfo{person}{Rui Li}, \bibinfo{person}{Xing Xie}, {and} \bibinfo{person}{Jian Tang}.} \bibinfo{year}{2022}\natexlab{}.
\newblock \showarticletitle{Learning on large-scale text-attributed graphs via variational inference}.
\newblock \bibinfo{journal}{\emph{arXiv preprint arXiv:2210.14709}} (\bibinfo{year}{2022}).
\newblock


\bibitem[Zhao et~al\mbox{.}(2023)]%
        {zhao2023learning}
\bibfield{author}{\bibinfo{person}{Jianan Zhao}, \bibinfo{person}{Meng Qu}, \bibinfo{person}{Chaozhuo Li}, \bibinfo{person}{Hao Yan}, \bibinfo{person}{Qian Liu}, \bibinfo{person}{Rui Li}, \bibinfo{person}{Xing Xie}, {and} \bibinfo{person}{Jian Tang}.} \bibinfo{year}{2023}\natexlab{}.
\newblock \bibinfo{title}{Learning on Large-scale Text-attributed Graphs via Variational Inference}.
\newblock
\showeprint[arxiv]{2210.14709}~[cs.LG]


\bibitem[Zhou et~al\mbox{.}(2009)]%
        {zhou2009predicting}
\bibfield{author}{\bibinfo{person}{Tao Zhou}, \bibinfo{person}{Linyuan L{\"u}}, {and} \bibinfo{person}{Yi-Cheng Zhang}.} \bibinfo{year}{2009}\natexlab{}.
\newblock \showarticletitle{Predicting missing links via local information}.
\newblock \bibinfo{journal}{\emph{The European Physical Journal B}}  \bibinfo{volume}{71} (\bibinfo{year}{2009}), \bibinfo{pages}{623--630}.
\newblock


\bibitem[Zhou et~al\mbox{.}(2022b)]%
        {zhou2022model}
\bibfield{author}{\bibinfo{person}{Xiao Zhou}, \bibinfo{person}{Yong Lin}, \bibinfo{person}{Renjie Pi}, \bibinfo{person}{Weizhong Zhang}, \bibinfo{person}{Renzhe Xu}, \bibinfo{person}{Peng Cui}, {and} \bibinfo{person}{Tong Zhang}.} \bibinfo{year}{2022}\natexlab{b}.
\newblock \showarticletitle{Model agnostic sample reweighting for out-of-distribution learning}. In \bibinfo{booktitle}{\emph{International Conference on Machine Learning}}. PMLR, \bibinfo{pages}{27203--27221}.
\newblock


\bibitem[Zhou et~al\mbox{.}(2022a)]%
        {zhou2022ood}
\bibfield{author}{\bibinfo{person}{Yangze Zhou}, \bibinfo{person}{Gitta Kutyniok}, {and} \bibinfo{person}{Bruno Ribeiro}.} \bibinfo{year}{2022}\natexlab{a}.
\newblock \showarticletitle{OOD link prediction generalization capabilities of message-passing GNNs in larger test graphs}.
\newblock \bibinfo{journal}{\emph{Advances in Neural Information Processing Systems}}  \bibinfo{volume}{35} (\bibinfo{year}{2022}), \bibinfo{pages}{20257--20272}.
\newblock


\bibitem[Zhu et~al\mbox{.}(2024)]%
        {zhu2024pitfalls}
\bibfield{author}{\bibinfo{person}{Jing Zhu}, \bibinfo{person}{Yuhang Zhou}, \bibinfo{person}{Vassilis~N Ioannidis}, \bibinfo{person}{Shengyi Qian}, \bibinfo{person}{Wei Ai}, \bibinfo{person}{Xiang Song}, {and} \bibinfo{person}{Danai Koutra}.} \bibinfo{year}{2024}\natexlab{}.
\newblock \showarticletitle{Pitfalls in Link Prediction with Graph Neural Networks: Understanding the Impact of Target-link Inclusion \& Better Practices}. In \bibinfo{booktitle}{\emph{Proceedings of the 17th ACM International Conference on Web Search and Data Mining}}. \bibinfo{pages}{994--1002}.
\newblock


\bibitem[Zhu et~al\mbox{.}(2021)]%
        {zhu2021neural}
\bibfield{author}{\bibinfo{person}{Zhaocheng Zhu}, \bibinfo{person}{Zuobai Zhang}, \bibinfo{person}{Louis-Pascal Xhonneux}, {and} \bibinfo{person}{Jian Tang}.} \bibinfo{year}{2021}\natexlab{}.
\newblock \showarticletitle{Neural bellman-ford networks: A general graph neural network framework for link prediction}.
\newblock \bibinfo{journal}{\emph{Advances in Neural Information Processing Systems}}  \bibinfo{volume}{34} (\bibinfo{year}{2021}), \bibinfo{pages}{29476--29490}.
\newblock


\end{thebibliography}

\appendix

\section{Additional Training Details} \label{sec:app_train_details}

This section provides relevant details about training and reproducing results not mentioned in Section~\ref{sec:experimental_setup}:
\begin{itemize}
    \item Please consult the project \href{https://github.com/revolins/LPShift}{README} for building the project, loading data, and re-creating results.
    \item All models and link predictors are fixed to 3 layers in their neural architectures.
    \item GCN and LPFormer were tuned to 128 hidden dimensions.
    \item BUDDY, NCNC, NeoGNN, and SEAL were tuned to 256 hidden dimensions.
    \item All experiments were conducted with a single A6000 48GB GPU and 1TB of available system RAM.
    \item We apply the 'NCNC2' variant of NCNC with an added \href{https://github.com/GraphPKU/NeuralCommonNeighbor/blob/main/README.md}{depth} argument of 2 \cite{wang2023neural} for all CN and SP splits of ogbl-collab.
    \item Otherwise, we set depth equal to 1, so as to reduce runtime.
    \item NeoGNN use 1-hop neighborhoods on all tested datasets.
    \item The learning rate and dropout for initial batch size tuning was fixed at $1e^{-3}$ and $0.1$ respectively.
    \item The model performance and memory complexity was tested in single runs with the following batch sizes: $\{8, 16, 32, 64, 128,$
    $256, 512, 1024, 2048,4096,8192, 16384, 32768,$
$ 65536 \}$.
    \item PPA Training Edges are limited to 3 million edges total.
    \item Validation and testing edges that are duplicated with training edges are removed from the edge index.
    \item In order to provide overlap within a given dataset, validation and testing edges that do not connect to training nodes are removed from the edge index.
    \item After sampling the necessary training edges, the adjacency matrix is extracted from the edge index, converted to an undirected graph and has any edge weights standardized to 1.
\end{itemize}



Common Neighbors, Preferential-Attachment and Shortest-Path, as respectively shown in Figure~\ref{fig:cn_split_ogb}, are interchangeable within the dataset splitting strategy. Details about how Common Neighbors functions within the strategy are included in Section~\ref{sec:d_splt_strat_dc}.

As another possible component of LPShift, Preferential-Attachment determines the degrees between a given source and target node and then multiples the two to produce the score, based on that score, the sample is then sorted into a new dataset split. 

As another possible component of LPShift, Shortest-Path determines the score by determining the minimum number of nodes necessary to reach the target node from the source node. If there is a link between the two nodes, we remove the link and then re-add to the adjacency matrix after the score calculation. The final Shortest-Path score applies the calculated shortest-path length, $SP(u,v)$ as the denominator in a ratio of $\frac{1}{SP(u,v)}$, which is then used to sort the sample into it's respective dataset split. 

\section{Cosine Similarity of ogbl-collab} \label{sec:app_analysis_details}
Given LPShift's effect on structure, we also quantify the extent to which it can affect the feature distributions of the real-world features within the ogbl-collab data. \textit{Note:} ogbl-ddi was not considered due to a lack of node features and ogbl-ppa consists of one-hot encoded features derived directly from structure. 

\begin{table}[h]
\centering
\caption{The cosine-similarity table of ogbl-collab's original feature distribution and its LPShift versions.}
\begin{tabular}{cc|l|l|l}
\toprule
\multicolumn{2}{c|}{Split}  & Train/Test & Train/Valid & Valid/Test \\ 
\midrule
&Original & 83.50 ± 7.33 & 83.40 ± 7.33 & 86.91 ± 6.58 \\
\midrule
\multirow{6}{*}{CN} &(0,1,2) & 87.73 ± 6.12 & 87.42 ± 6.17 & 91.78 ± 4.01 \\
&(0,2,4) & 87.06 ± 6.08 & 86.72 ± 6.14 & 91.01 ± 4.29 \\
&(0,3,5) & 86.45 ± 6.12 & 86.19 ± 6.12 & 90.42 ± 4.41 \\
  \cmidrule{2-5}
&(2,1,0) & 85.53 ± 6.74 & 85.22 ± 6.79 & 90.45 ± 4.49 \\
&(4,2,0) & 85.97 ± 7.09 & 85.55 ± 7.15 & 90.71 ± 4.43 \\
&(5,3,0) & 86.08 ± 7.29 & 85.62 ± 7.34 & 90.74 ± 4.53 \\
\midrule
\multirow{4}{*}{SP} &($\infty$,6,4) & 84.11 ± 6.94 & 82.12 ± 7.34 & 85.60 ± 6.75 \\
&($\infty$,4,3) & 84.35 ± 6.93 & 82.66 ± 7.45 & 85.97 ± 6.75 \\
  \cmidrule{2-5}
&(4,6,$\infty$) & 81.10 ± 7.74 & 82.15 ± 7.79 & 81.86 ± 7.48 \\
&(3,4,$\infty$) & 81.99 ± 7.84 & 82.81 ± 7.98 & 82.76 ± 7.72 \\
\midrule
\multirow{6}{*}{PA} &(0,50,100) & 83.92 ± 7.22 & 82.59 ± 7.63 & 85.50 ± 7.14 \\
&(0,100,200) & 87.48 ± 6.16 & 87.55 ± 6.10 & 92.52 ± 3.70 \\
&(0,150,250) & 87.61 ± 6.10 & 87.81 ± 6.17 & 92.45 ± 3.84 \\
  \cmidrule{2-5}
&(100,50,0) & 82.28 ± 7.96 & 83.22 ± 8.00 & 83.17 ± 7.85 \\
&(200,100,0) & 84.75 ± 6.65 & 85.32 ± 6.64 & 89.94 ± 4.52 \\
&(250,150,0) & 85.08 ± 6.65 & 85.51 ± 6.64 & 90.62 ± 4.46 \\
\bottomrule
\end{tabular}
\label{table:cosine_similarity}
\end{table}

\section{Time to Split Tested Dataset Samples} \label{sec:avg_time_tests}

A key consideration for LPShift's application as a splitting strategy is to alleviate the burden of gathering new datasets, allowing researchers to control for and then induce a distribution shift in link-prediction datasets quickly; without requiring an expensive and time-consuming project to build a new dataset. This consideration is inspired by current graph and node-classification benchmark datasets, all of which induce distribution shifts in pre-existing benchmark datasets~\cite{gui2022good},\cite{ji2022drugood},\cite{koh2021wilds}. LPShift is not meant to replace high-quality benchmark datasets, especially for distribution shifts, but to serve as a supplement for current datasets and enhance understanding of LP generalization. Results demonstrating LPShift's time-efficiency on tested dataset splits are included below in Table~\ref{table:time_to_gen}.

\begin{table}[h]
\centering
\caption{The average time in seconds (s) across 10 runs to generate each 'Forward' and 'Backward' split for ogbl-ppa, ogbl-collab, ogbl-ddi.}
\begin{tabular}{cc|ccc}
\toprule
\multicolumn{2}{c|}{Split} & ogbl-collab & ogbl-ppa & ogbl-ddi \\
\midrule
 \multirow{6}{*}{CN} & (0, 1, 2)  & 7.48 s  & 177.89 s &  13.45 s \\
  &(0, 2, 4)  & 7.49 s  & 177.09 s & 14.21 s \\
  &(0, 3, 5)  & 7.63 s  & 178.23 s & 13.64 s \\
  \cmidrule{2-5}
  &(2, 1, 0)    & 7.66 s  & 184.98 s & 13.31 s \\
&(4, 2, 0)    & 7.63 s  & 186.49 s & 13.71 s\\
&(5, 3, 0)    & 7.86 s  & 185.95 s & 13.42 s\\
\midrule
\multirow{4}{*}{SP} & ($\infty$, 6, 4)  & 53.12 s & 2748.64 s & -- \\
 &($\infty$, 4, 3)  & 52.24 s & 2705.91 s & -- \\
 \cmidrule{2-5}
 &(4, 6, $\infty$) & 53.93 s & 2715.34 s & -- \\
&(3, 4, $\infty$) & 53.7 s  & 2751.56 s & -- \\
\midrule
\multirow{6}{*}{PA} & (0, 50, 100) & 19.25 s & 406.5 s & 15.41 s \\
 &(0, 100, 200) & 19.05 s & 408.04 s & 14.55 s\\
 &(0, 150, 250) & 19.35 s & 407.81 s & 13.96 s\\
 \cmidrule{2-5}
&(100, 50, 0) & 19.65 s & 425.3 s & 14.42 s\\
&(200, 100, 0) & 19.34 s & 409.93 s & 14.59 s\\
&(250, 150, 0) & 19.37 s & 403.55 s & 14.05 s\\
\bottomrule
\end{tabular}
\label{table:time_to_gen}
\end{table}

\section{Size of Dataset Samples} \label{sec:app_generalization_results}
In order to aid reproducibility of results when running LPShift manually; this section details the number of training, validation, and test edges for all LPShift dataset split types introduced within in Section~\ref{sec:create_datasets} and all datasets.

\begin{table}[h]
\centering
\caption{The number of ogbl-collab samples for the forward and backward heuristic splits.}
\begin{adjustbox}{max width=\linewidth}
\begin{tabular}{c|cccccc}
\toprule
CN & (0,1,2) & (0,2,4) & (0,3,5) & (2,1,0) & (4,2,0) & (5,3,0) \\
\midrule
Train & 57638 & 237928 & 493790 & 1697336 & 1193456 & 985820 \\
Valid & 6920  & 20045  & 31676  & 23669   & 24097   & 25261 \\
Test  & 4326  & 14143  & 21555  & 9048    & 11551   & 11760 \\
\midrule
SP & ($\infty$,6,4) & ($\infty$,4,3) & (4,6,$\infty$) & (3,4,$\infty$) & -- & --\\
\midrule
Train & 46880 & 52872 & 1882392 & 1877626 & -- & --\\
Valid & 1026  & 1238  & 5222    & 4384    & -- & --\\
Test  & 2759  & 3457  & 2626    & 7828    & -- & --\\
\midrule
PA & (0,50,100) & (0,100,200) & (0,150,250) & (100,50,0) & (200,100,0) & (250,150,0) \\
\midrule
Train & 210465 & 329383 & 409729 & 1882392 & 1877626 & 457372 \\
Valid & 46626  & 62868  & 75980  & 64729   & 64202   & 65323 \\
Test  & 9492   & 25527  & 39429  & 41381   & 30983   & 30999 \\
\bottomrule
\end{tabular}
\end{adjustbox}
\label{table:collab_samp_size}
\end{table}



\begin{table}[h]
\centering
\caption{The number of ogbl-ppa samples for the forward and backward heuristic splits.}
\begin{adjustbox}{max width=\linewidth}
\begin{tabular}{c|cccccc}
\toprule
CN & (0,1,2) & (0,2,4) & (0,3,5) & (2,1,0) & (4,2,0) & (5,3,0) \\
\midrule
Train & 2325936 & 3000000 & 3000000 & 3000000 & 3000000 & 3000000 \\
Valid & 87880   & 95679   & 98081   & 96765   & 93210   & 92403 \\
Test  & 67176   & 83198   & 88778   & 92798   & 85448   & 81887 \\
\midrule
SP & ($\infty$,6,4) & ($\infty$,4,3) & (4,$\infty$,6) & (3,4,$\infty$) & --& --\\
\midrule
Train & 17464 & 134728 & 3000000 & 3000000 & --& --\\
Valid & 149   & 4196   & 90511   & 97121   & --& --\\
Test  & 20    & 1180   & 458     & 74068   & --& --\\
\midrule
PA & (0,5k,10k) & (0,10k,20k) & (0,15k,25k) & (10k,5k,0) & (20k,10k,0) & (25k,15k,0) \\
\midrule
Train & 95671   & 98562   & 99352   & 3000000 & 3000000 & 3000000 \\
Valid & 95671   & 98562   & 99352   & 90623   & 89671   & 91995 \\
Test  & 45251   & 63178   & 72382   & 44593   & 34321   & 35088 \\
\bottomrule
\end{tabular}
\end{adjustbox}
\label{table:ppa_samp_size}
\end{table}


\begin{table}[h]
\centering
\small
\caption{The number of ogbl-ddi samples for the forward and backward heuristic splits.}
\begin{adjustbox}{max width=\linewidth}
\begin{tabular}{c|cccccc}
\toprule
CN & (0,1,2) & (0,2,4) & (0,3,5) & (2,1,0) & (4,2,0) & (5,3,0) \\
\midrule
Train & 3938  & 5636  & 6580  & 10000 & 10000 & 10000 \\
Valid & 1184  & 992   & 942   & 966   & 1668  & 1462 \\
Test  & 212   & 525   & 809   & 1808  & 2296  & 2832 \\
\midrule
PA & (0,5k,10k) & (0,10k,20k) & (0,15k,25k) & (10k,5k,0) & (20k,10k,0) & (25k,15k,0) \\
\midrule
Train & 5132  & 8246  & 11718 & 10000 & 10000 & 10000 \\
Valid & 1572  & 3208  & 5058  & 2678  & 6188  & 6300 \\
Test  & 419   & 1422  & 3694  & 1900  & 3632  & 5970 \\
\bottomrule
\end{tabular}
\end{adjustbox}
\label{table:ddi_samp_size}
\end{table}

\section{Is LPShift effective at inducing distribution shift?} \label{sec:app_lpshift_effect}

The following section will explore the capability of LPShift to induce a measurably-significant distribution shift in the structure of the ogbl-collab dataset. We apply the 2-sample Kolmgorov-Smirnov (KS) test~\cite{hodges1958significance} to compare if training, validation, and testing distributions can be sampled from one another, both before and after applying LPShift. As a controlled baseline to test LPShift, we generate a Holme-Kim (HK) graph~\cite{holme2002growing} with a 40\% chance to close a triangle, allowing the HK graph to contain numerous Common Neighbors without becoming fully-connected. HK graph generation parameters are included in Appendix~\ref{sec:app_analysis_details}.

\begin{figure}[h]
    \centering
    \includegraphics[width=\linewidth]{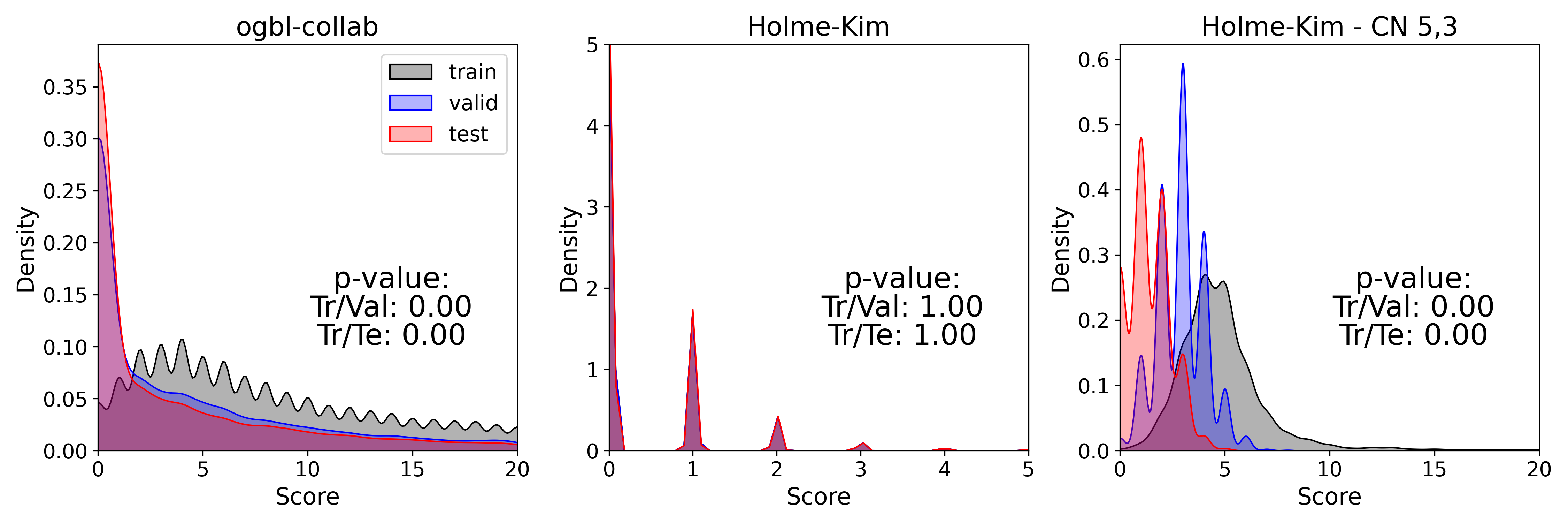}
    \caption{Three subplots detailing CN distributions for: 1.) the unaltered ogbl-collab dataset 2.) a Holme-Kim (HK) graph with a random split 3.) the HK graph from 2. split with LPShift's CN - 5,3 strategy}
    \Description{Three subplots detailing CN distributions for: 1.) the unaltered ogbl-collab dataset 2.) a Holme-Kim (HK) graph with a random split 3.) the HK graph from 2. split with LPShift's CN - 5,3 strategy}
    \label{fig:collab_vs_synth_cn}
\end{figure}

The Holme-Kim graph used for analysis in Figure~\ref{fig:collab_vs_synth_cn} was generated with the following parameters:
\begin{itemize}
    \item $n$ = $235868$, $m$ = $5$, $p_{c}$ = $0.4$, seed = $42$ 
\end{itemize}

The first subplot in Figure~\ref{fig:collab_vs_synth_cn} extends the reasoning introduced with NCNC~\cite{wang2023neural}. As such, this subplot indicates there is a natural shift for CNs within the original ogbl-collab dataset \cite{hu2020open}. The p-value of 0, measured across both split permutations, indicates that the training distribution of CNs is shifted from the validation and testing distributions. The second subplot depicts a randomly-split HK graph~\cite{holme2002growing}, where CN distributions for each split match one another, further indicated by the p-values of 1. The third subplot depicts the HK graph from the second subplot split with LPShift's CN - 5,3 strategy, resulting in a distinct shift between all dataset splits, as confirmed by the 0 p-values. As such, LPShift induces structural shift that is as measurably dissimilar as the structural shift present in the original ogbl-collab dataset, even when the initial dataset splits are measurably identical. Additionally, the CN - 5,3 split causes the shape of the HK graph's CN distributions to become more similar to the shift observed within the original ogbl-collab dataset, indicating that the "Backward" LPShift strategy can function like a real-world distribution shift. 

\section{Dataset Results} \label{sec:app_dataset_res_tables}

This section includes~\Crefrange{table:collab_mrr}{table:ddi_mrr}, displaying baseline results for initial LPShift experiments.

\begin{table}[h]
\centering
\caption{{\bf ogbl-collab} results reported in MRR with the best {\bf bolded} and the second best \underline{underlined}.}
\begin{adjustbox}{max width=\linewidth}
\begin{tabular}{cc|c|c|c|c|c|c|c}
\toprule
\multicolumn{2}{c|}{\multirow{2}{*}{Split}} & \multicolumn{7}{c}{Models} \\ 
&& RA & GCN & BUDDY & NCNC & LPFormer & NeoGNN & SEAL \\ 
\midrule
\multirow{6}{*}{CN} & (0, 1, 2) & {\bf 32.22} & 14.89 ± 0.67 & \underline{17.48 ± 1.19} & 5.34 ± 2.54 & 4.27 ± 1.17 & 5.53 ± 1.52 & 7.06 ± 1.67 \\ 
&(0, 2, 4) & {\bf 29.74} & 17.75 ± 0.44 & 15.47 ± 0.57 & 13.99 ± 1.35 & 12.97 ± 1.24 & 13.26 ± 0.39 & \underline{20.60 ± 6.23} \\ 
&(0, 3, 5) & {\bf 29.86} & 19.08 ± 0.26 & 16.60 ± 0.89 & 15.06 ± 1.43 & \underline{25.36 ± 2.04} & 14.21 ± 0.88 & 19.78 ± 1.24 \\ 
\cmidrule{2-9}
 &(2, 1, 0) & 0.6 & \textbf{5.04 ± 0.13} & \underline{3.70 ± 0.13} & 1.46 ± 0.02 & 2.01 ± 0.95 & 2.16 ± 0.05 & 1.01 ± 0.02 \\
&(4, 2, 0) & 4.79 & \underline{4.87 ± 0.17} & 3.55 ± 0.09 & \textbf{6.13 ± 0.49} & 3.87 ± 0.74 & 3.41 ± 0.16 & 3.38 ± 0.62 \\
& (5, 3, 0) & \textbf{15.9} & 6.78 ± 0.19 & 5.40 ± 0.04 & \underline{12.87 ± 1.50} & 8.36 ± 0.63 & 7.85 ± 0.74 & 8.08 ± 1.96 \\
\midrule
\multirow{4}{*}{SP} & ($\infty$, 6, 4) & {\bf 33.87} & 12.64 ± 0.85 & \underline{16.20 ± 1.40} & 14.43 ± 1.36 & 4.6 ± 3.15 & 6.14 ± 1.44 & 2.08 ± 1.50 \\ 
& ($\infty$, 4, 3) & {\bf 33.91} & 15.50 ± 0.16 & 16.42 ± 2.30 & \underline{18.33 ± 1.24} & 15.7 ± 2.87 & 5.82 ± 1.20 & 2.10 ± 1.24 \\ 
\cmidrule{2-9}
 & (4, 6, $\infty$) & 0.69 & \underline{4.75 ± 0.10} & \textbf{6.73 ± 0.32} & 0.73 ± 0.13 & 3.16 ± 0.62 & 4.55 ± 0.27 & 0.93 ± 0.07 \\
& (3, 4, $\infty$) & 0.63 & \textbf{3.81 ± 0.16} & \underline{3.71 ± 0.39} & 0.95 ± 0.24 & 1.86 ± 0.48 & 2.54 ± 0.09 & 0.80 ± 0.01 \\
\midrule
\multirow{6}{*}{PA} & (0, 50, 100) & {\bf 36.87} & 19.81 ± 0.59 & 21.27 ± 0.74 & 17.24 ± 0.17 & 25.31 ± 5.67 & 15.94 ± 0.49 & \underline{29.06 ± 1.57} \\ 
&(0, 100, 200) & {\bf 26.78} & 12.68 ± 0.20 & 14.04 ± 0.76 & 12.34 ± 3.01 & 11.98 ± 3.12 & 13.50 ± 0.84 & \underline{20.69 ± 1.51} \\ 
&(0, 150, 250) & {\bf 24.07} & 12.73 ± 0.32 & 13.06 ± 0.53 & 12.01 ± 0.82 & 12.43 ± 6.62 & 12.81 ± 0.51 & \underline{16.23 ± 3.69} \\
\cmidrule{2-9}
 & (100, 50, 0) & \textbf{33.09} & 24.33 ± 0.537 & 24.95 ± 0.92 & 20.81 ± 1.86& 17.76 ± 2.01 & 6.42 ± 0.462 & \underline{31.83 ± 6.44} \\
& (200, 100, 0) & \textbf{42.28} & 13.97 ± 0.52 & 15.52 ± 0.73 & 18.62 ± 1.54 & 27.56 ± 9.10 & 4.20 ± 0.37 & \underline{31.96 ± 6.65} \\
& (250, 150, 0) & \textbf{44.14} & 12.98 ± 0.45 & 13.36 ± 0.83 & 18.71 ± 0.99 & 24.04 ± 11.35 & 4.40 ± 0.63 & \underline{39.58 ± 4.81} \\
\bottomrule
\end{tabular}
\label{table:collab_mrr}
\end{adjustbox}
\end{table}

\begin{table}[h]
\centering
\caption{{\bf ogbl-ppa} results reported in MRR with the best {\bf bolded} and the second best \underline{underlined}.}
\begin{adjustbox}{max width=\linewidth}
\begin{tabular}{cc|c|c|c|c|c|c|c}
\toprule
\multicolumn{2}{c|}{\multirow{2}{*}{Split}} & \multicolumn{7}{c}{Models} \\ 
&& RA & GCN & BUDDY & NCNC & LPFormer & NeoGNN & SEAL \\ 
\midrule
\multirow{6}{*}{CN} & (0,1,2) & 4.71 & 7.81 ± 0.08 & \underline{7.90 ± 0.32} & 7.08 ± 3.08 & 3.28 ± 0.63 & 3.49 ± 0.10 & \textbf{11.91 ± 1.85} \\
& (0,2,4) & 4.45 & \underline{8.18 ± 0.12} & 3.83 ± 0.24 & \textbf{9.89 ± 1.92} & 2.46 ± 0.51 & 5.79 ± 0.42 & 4.84 ± 0.10 \\
& (0,3,5) & 4.38 & \textbf{8.75 ± 0.19} & 3.06 ± 0.06 & \underline{8.22 ± 1.15} & 4.84 ± 0.73 & 5.71 ± 0.26 & 5.15 ± 0.10 \\
\cmidrule{2-9}
& (2,1,0) & 0.53 & \underline{2.59 ± 0.08} & 1.60 ± 0.05 & 2.37 ± 0.15 & \textbf{6.04 ± 0.41} & 0.76 ± 0.02 & 1.03 ± 0.54 \\
& (4,2,0) & 0.92 & 2.16 ± 0.05 & 2.47 ± 0.07 & \textbf{8.54 ± 0.74} & \underline{4.23 ± 0.46} & 0.79 ± 0.00 & 0.95 ± 0.09 \\
& (5,3,0) & 1.17 & 2.17 ± 0.04 & 2.56 ± 0.08 & \textbf{9.04 ± 0.92} & \underline{3.87 ± 0.10} & 0.86 ± 0.02 & 1.35 ± 0.56 \\
\midrule
\multirow{4}{*}{SP} & ($\infty$,6,4) & \textbf{32.57} & 9.95 ± 0.52 & 1.24 ± 0.02 & \underline{11.16 ± 8.81} & 9.83 ± 5.92 & 3.13 ± 0.38 & 11.14 ± 12.06 \\
& ($\infty$,4,3) & \textbf{19.84} & 3.29 ± 0.10 & \underline{5.87 ± 0.16} & 0.79 ± 0.54 & 4.94 ± 0.62 & 3.58 ± 0.45 & 2.96 ± 4.58 \\
\cmidrule{2-9}
& (4,6,$\infty$) & 0.65 & \textbf{10.53 ± 0.48} & \underline{9.95 ± 0.52} & 5.84 ± 0.34 & 5.90 ± 1.76 & 4.89 ± 0.13 & 1.51 ± 0.72 \\
& (3,4,$\infty$) & 0.54 & \underline{3.38 ± 0.11} & \textbf{5.87 ± 0.16} & 1.01 ± 0.19 & 1.38 ± 0.46 & 0.83 ± 0.01 & 0.51 ± 0.02 \\
\midrule
\multirow{6}{*}{PA} & (0,5k,10k) & 3.9  & 3.69 ± 0.76 & 3.93 ± 0.98 & \underline{8.00 ± 0.60} & \textbf{9.27 ± 1.78} & 4.92 ± 0.58 & 4.22 ± 0.63 \\
& (0,10k,20k) & 3.14 & 3.42 ± 0.59 & 6.38 ± 3.48 & \underline{6.90 ± 1.46} & \textbf{9.03 ± 1.6}  & 6.29 ± 0.87 & 3.43 ± 0.19 \\
& (0,15k,25k) & 2.72 & 5.62 ± 0.40 & 2.48 ± 0.03 & 8.00 ± 0.78 & \textbf{9.07 ± 2.43} & \underline{8.98 ± 1.10} & 3.57 ± 0.74 \\
\cmidrule{2-9}
& (10k,5k,0) & 7.4 & 3.09 ± 0.15 & 3.15 ± 0.16 & \underline{10.56 ± 0.73} & \textbf{14.43 ± 4.45} & 1.52 ± 0.05 & 4.88 ± 0.90 \\
& (20k,10k,0) & 5.81 & 2.50 ± 0.23 & 2.55 ± 0.16 & \textbf{7.74 ± 0.46} & \underline{8.43 ± 3.46} & 1.38 ± 0.04 & 4.50 ± 1.10 \\
& (25k,15k,0) & 5.08 & 2.79 ± 0.28 & 2.37 ± 0.02 & \textbf{7.82 ± 0.36} & \underline{6.27 ± 3.87} & 1.39 ± 0.06 & 2.38 ± 0.73 \\
\bottomrule
\end{tabular}
\end{adjustbox}
\label{table:ppa_mrr}
\end{table}

\clearpage

\begin{table}[h]
\centering
\caption{{\bf ogbl-ddi} results reported in MRR with the best {\bf bolded} and the second best \underline{underlined}.}
\begin{adjustbox}{max width=\linewidth}
\begin{tabular}{cc|c|c|c|c|c|c|c}
\toprule
\multicolumn{2}{c|}{\multirow{2}{*}{Split}} & \multicolumn{7}{c}{Models} \\ 
&& RA & GCN & BUDDY & NCNC & LPFormer & NeoGNN & SEAL \\ 
\midrule
\multirow{6}{*}{CN} & (0,1,2) & \underline{4.8} &\textbf{5.34 ± 0.24} &1.65 ± 0.17 &0.79 ± 0.20 & 0.82 ± 0.21& 0.92 ± 0.01& 0.89 ± 0.39\\
& (0,2,4) & \underline{4.25} & \textbf{4.88 ± 0.41}& 2.75 ± 0.16&1.39 ± 0.81 &0.85 ± 0.14 &2.12 ± 0.77 &1.17 ± 0.32\\
& (0,3,5) & \textbf{3.86} &\underline{3.48 ± 0.26} &2.15 ± 0.08 & 1.88 ± 0.57& 1.06 ± 0.17&1.75 ± 0.44 & 0.98 ± 0.37\\
\cmidrule{2-9}
& (2,1,0) & 0.46 &1.14 ± 0.20 &0.78 ± 0.06 &\underline{2.05 ± 2.58} &\textbf{3.13 ± 0.63} &0.94 ± 0.40 &0.87 ± 0.49 \\
& (4,2,0) & 0.59  & 1.01 ± 0.20&1.09 ± 0.22 &0.61 ± 0.17 &\textbf{3.46 ± 0.61} &\underline{2.66 ± 1.94} & 0.62 ± 0.08\\
& (5,3,0) & 0.54 &1.41 ± 0.32 &1.40 ± 0.35 & 0.58 ± 0.03&\textbf{3.94 ± 0.28} &\underline{2.69 ± 1.89} & 0.62 ± 0.07\\
\midrule
\multirow{6}{*}{PA} & (0,5k,10k) & \textbf{5.29} & 2.61 ± 0.38&1.81 ± 0.35 & \underline{3.44 ± 1.73}&1.19 ± 0.11 &1.96 ± 0.72 & 0.83 ± 0.30\\
& (0,10k,20k) & \textbf{3.6}  &1.96 ± 0.22 &1.74 ± 0.28 &\underline{2.70 ± 0.53} &0.84 ± 0.21 &1.31 ± 0.30 & 1.35 ± 0.87\\
& (0,15k,25k) & \textbf{2.54} &1.91 ± 0.14 &1.92 ± 0.08 &\underline{2.33 ± 1.18} &1.46 ± 0.31 &1.31 ± 0.20 & 1.48 ± 0.69\\
\cmidrule{2-9}
& (10k,5k,0) &7.06  &4.68 ± 1.19 &1.19 ± 0.32 &\underline{12.55 ± 15.18} &\textbf{14.19 ± 5.59} &3.63 ± 0.50 & 4.64 ± 4.21\\
& (20k,10k,0) & 1.5 &2.04 ± 0.77 &2.71 ± 1.60 &\textbf{10.03 ± 9.20} &\underline{8.26 ± 1.13} &2.89 ± 0.81 & 1.31 ± 0.92\\
& (25k,15k,0) & 1.68 & 1.14 ± 0.39&2.09 ± 0.72 &\underline{4.66 ± 4.89} &\textbf{8.51 ± 0.82} &1.58 ± 0.98 & 1.71 ± 0.56\\
\bottomrule
\end{tabular}
\end{adjustbox}
\label{table:ddi_mrr}
\end{table}

\section{Generalization Results} \label{sec:generalization_perf}
This section includes~\Crefrange{tab:collab_lpgen}{tab:ddi_lpgen}, displaying results for the graph-specific generalization methods conducted on the LPShift dataset splits.

\begin{table}[h]
\centering
\caption{ogbl-collab (MRR) results under LPShift after applying graph-specific generalization methods.}
\begin{adjustbox}{max width=\linewidth}
\begin{tabular}{c|cccccc}
\toprule
CN & (0,1,2) & (0,2,4) & (0,3,5) & (2,1,0) & (4,2,0) & (5,3,0) \\
\midrule
DropEdge     & 15.54 ± 0.98 & 16.16 ± 0.17 & 16.34 ± 0.17 & 2.61 ± 0.09  & 2.88 ± 0.11 & 4.99 ± 0.08 \\
TC           & 11.27 ± 2.03 & 12.39 ± 1.43 & 14.99 ± 1.69 & 5.28 ± 0.05  & 5.23 ± 0.05 & 6.03 ± 0.10 \\
GCN+BUDDY    & 8.50 ± 1.10  & 12.85 ± 0.83 & 15.35 ± 0.96 & 5.46 ± 0.11  & 5.36 ± 0.12 & 5.96 ± 0.06 \\
RA+BUDDY     & 3.94 ± 0.51  & 6.61 ± 0.30  & 7.16 ± 0.09  & 4.32 ± 0.19  & 4.45 ± 0.12 & 4.93 ± 0.11 \\
\midrule
SP & ($\infty$,6,4) & ($\infty$,4,3) & (4,6,$\infty$) & (3,4,$\infty$) & --& --\\
\midrule
DropEdge     & 12.31 ± 0.51 & 17.11 ± 1.02 & 5.34 ± 0.43  & 2.93 ± 0.18 & --& --\\
TC           & 7.25 ± 0.81  & 9.33 ± 1.66  & 6.88 ± 0.30  & 6.24 ± 0.13 & --& --\\
GCN+BUDDY    & 7.38 ± 0.82  & 9.60 ± 0.39  & 6.86 ± 1.13  & 6.47 ± 0.24 & --& --\\
RA+BUDDY     & 3.84 ± 0.59  & 7.63 ± 0.25  & 6.86 ± 1.13  & 6.47 ± 0.24 & --& --\\
\midrule
PA & (0,50,100) & (0,100,200) & (0,150,250) & (100,50,0) & (200,100,0) & (250,150,0) \\
\midrule
DropEdge     & 21.35 ± 0.36 & 13.84 ± 0.64 & 12.85 ± 0.78 & 26.09 ± 0.62 & 15.68 ± 0.85 & 13.13 ± 0.94 \\
TC           & 18.82 ± 1.35 & 12.13 ± 1.04 & 10.63 ± 0.55 & 15.84 ± 1.13 & 9.15 ± 0.39  & 6.70 ± 0.20 \\
GCN+BUDDY    & 15.92 ± 1.01 & 9.47 ± 0.31  & 9.60 ± 0.41  & 14.34 ± 1.05 & 8.35 ± 0.34  & 5.50 ± 0.33 \\
RA+BUDDY     & 15.92 ± 1.01 & 9.47 ± 0.31  & 9.60 ± 0.41  & 14.34 ± 1.05 & 8.35 ± 0.34  & 5.50 ± 0.33 \\
\bottomrule
\end{tabular}
\end{adjustbox}
\label{tab:collab_lpgen}
\end{table}

\begin{table}[h]
\centering
\caption{ogbl-ppa (MRR) results under LPShift after applying graph-specific generalization methods.}
\begin{adjustbox}{max width=\linewidth}
\begin{tabular}{c|cccccc}
\toprule
CN& (0,1,2) & (0,2,4) & (0,3,5) & (2,1,0) & (4,2,0) & (5,3,0) \\
\midrule
DropEdge     & 7.83 ± 0.27 & 3.83 ± 0.25 & 3.06 ± 0.06 & 1.61 ± 0.04 & 2.47 ± 0.07 & 2.56 ± 0.08 \\
TC           & 5.27 ± 0.34 & 2.91 ± 0.06 & 2.67 ± 0.13 & 3.44 ± 0.08 & 3.45 ± 0.10 & 3.55 ± 0.13 \\
GCN+BUDDY    & 4.48 ± 0.33 & 3.79 ± 0.28 & 3.16 ± 0.10 & 3.19 ± 0.08 & 3.25 ± 0.09 & 3.36 ± 0.13 \\
RA+BUDDY     & 4.04 ± 0.26 & 3.42 ± 0.20 & 2.95 ± 0.13 & 2.58 ± 0.09 & 3.04 ± 0.05 & 3.10 ± 0.15 \\
\midrule
SP& ($\infty$,6,4) & ($\infty$,4,3) & (4,6,$\infty$) & (3,4,$\infty$) & --& --\\
\midrule
DropEdge     & 3.86 ± 0.39 & 5.87 ± 0.16 & 3.86 ± 0.39 & 5.87 ± 0.16 & --& --\\
TC           & 4.00 ± 0.29 & 4.82 ± 0.39 & 13.2 ± 0.45 & 2.89 ± 0.09 & --& --\\
GCN+BUDDY    & 4.00 ± 0.20 & 5.53 ± 0.92 & 14.41 ± 0.67 & 2.93 ± 0.15 & --& --\\
RA+BUDDY     & 3.89 ± 0.23 & 5.53 ± 0.94 & 13.63 ± 0.97 & 2.51 ± 0.14 & --& --\\
\midrule
PA& (0,5k,10k) & (0,10k,20k) & (0,15k,25k) & (10k,5k,0) & (20k,10k,0) & (25k,15k,0) \\
\midrule
DropEdge     & 3.93 ± 0.98 & 6.38 ± 3.48 & 2.49 ± 0.01 & 3.13 ± 0.10 & 2.56 ± 0.19 & 2.40 ± 0.03 \\
TC           & 3.62 ± 0.21 & 3.13 ± 0.12 & 3.33 ± 1.38 & 1.78 ± 0.16 & 1.50 ± 0.0008 & 1.53 ± 0.04 \\
GCN+BUDDY    & 3.66 ± 0.38 & 3.19 ± 0.22 & 2.88 ± 0.06 & 2.05 ± 0.05 & 1.67 ± 0.03 & 1.66 ± 0.02 \\
RA+BUDDY     & 3.66 ± 0.38 & 3.19 ± 0.22 & 2.88 ± 0.06 & 2.05 ± 0.04 & 1.65 ± 0.05 & 1.64 ± 0.03 \\
\bottomrule
\end{tabular}
\end{adjustbox}
\label{tab:ppa_lpgen}
\end{table}


\begin{table}[h]
\centering
\small
\caption{ogbl-ddi (MRR) results under LPShift after applying graph-specific generalization methods.}
\begin{adjustbox}{max width=\linewidth}
\begin{tabular}{c|cccccc}
\toprule
\multicolumn{7}{c}{\textbf{CN}} \\
& (0,1,2) & (0,2,4) & (0,3,5) & (2,1,0) & (4,2,0) & (5,3,0) \\
\midrule
DropEdge     & 1.65 ± 0.17 & 2.75 ± 0.16 & 2.15 ± 0.08 & 0.78 ± 0.06 & 1.09 ± 0.22 & 1.40 ± 0.35 \\
TC           & 1.10 ± 0.01 & 1.01 ± 0.01 & 0.94 ± 0.01 & 1.04 ± 0.02 & 0.90 ± 0.02 & 0.98 ± 0.03 \\
GCN+BUDDY    & 0.87 ± 0.21 & 1.45 ± 0.35 & 1.65 ± 0.22 & 1.20 ± 0.14 & 2.44 ± 0.27 & 3.19 ± 0.36 \\
RA+BUDDY     & 0.87 ± 0.21 & 1.45 ± 0.35 & 1.65 ± 0.22 & 1.20 ± 0.14 & 2.44 ± 0.27 & 3.19 ± 0.36 \\
\midrule
\multicolumn{7}{c}{\textbf{PA}} \\
& (0,5k,10k) & (0,10k,20k) & (0,15k,25k) & (10k,5k,0) & (20k,10k,0) & (25k,15k,0) \\
\midrule
DropEdge     & 1.81 ± 0.35 & 1.74 ± 0.28 & 1.92 ± 0.08 & 3.83 ± 3.22 & 2.71 ± 1.60 & 11.07 ± 1.44 \\
TC           & 1.04 ± 0.01 & 1.06 ± 0.02 & 1.24 ± 0.02 & 1.00 ± 0.03 & 1.66 ± 0.01 & 1.95 ± 0.02 \\
GCN+BUDDY    & 1.26 ± 0.28 & 1.43 ± 0.10 & 1.04 ± 0.23 & 9.16 ± 9.27 & 9.84 ± 1.97 & 9.50 ± 2.18 \\
RA+BUDDY     & 1.26 ± 0.28 & 1.43 ± 0.11 & 1.04 ± 0.23 & 9.16 ± 9.27 & 9.84 ± 1.97 & 9.50 ± 2.18 \\
\bottomrule
\end{tabular}
\end{adjustbox}
\label{tab:ddi_lpgen}
\end{table}
\pagebreak

\section{Traditional OOD Generalization Method Results} \label{sec:app_ood_gen_results}

This section includes table detailing the performance of domain-agnostic/traditional OOD generalization methods on all of LPShift's tested dataset splits.

\begin{table}[h]
\centering
\caption{MRR Results for traditional OOD generalization methods on LPShift's {\bf ogbl-collab}.}
\begin{adjustbox}{max width=\linewidth}
\begin{tabular}{l|ccc|cc|ccc}
\toprule
 \multirow{1}{*}{Models} & \multicolumn{3}{c}{CN Splits} &\multicolumn{2}{c}{SP Splits}  &\multicolumn{3}{c}{PA Splits}   \\ 
& (0, 1, 2) & (0, 2, 4) & (0, 3, 5) & ($\infty$, 6, 4) & ($\infty$, 4, 3) & (0, 50, 100) & (0, 100, 200) & (0, 150, 250) \\
  \midrule
  GCN & 14.89 ± 0.67&	17.75 ± 0.44	&19.08 ± 0.26	&12.64 ± 0.85	&15.50 ± 0.16	&19.81 ± 0.59	&12.68 ± 0.20	&12.73 ± 0.32\\
  \cmidrule(lr){2-9}
  +IRM &7.40 ± 0.32 &6.97 ± 0.55 &5.50 ± 0.71 & 5.84 ± 0.58&5.28 ± 0.17 & 11.18 ± 1.46&6.88 ± 1.65 &6.95 ± 1.09  \\
  +VREx &14.31 ± 0.76 &16.95 ± 0.22 &18.86 ± 0.13 &13.17 ± 0.58 &15.75 ± 0.50 &21.69 ± 0.18 &15.23 ± 0.44 & 14.66 ± 0.14 \\
  +GroupDRO &5.13 ± 0.55  &5.19 ± 0.61 &5.51 ± 0.67 &6.27 ± 0.91 & 6.36 ± 0.53&9.56 ± 1.93 &7.39 ± 0.65 & 6.04 ± 0.90  \\
  +DANN &15.69 ± 0.28  &16.83 ± 0.18 &18.80 ± 0.20 &14.01 ± 0.37 &15.60 ± 0.45 &21.77 ± 0.31 &15.61 ± 0.30 &  14.89 ± 0.11 \\
  +Deep CORAL &14.31 ± 0.76  &16.95 ± 0.22 &18.84 ± 0.10 &13.26 ± 0.68 &15.60 ± 0.29 &21.69 ± 0.18 & 15.00 ± 0.28& 14.59 ± 0.13  \\
 \midrule
& (2, 1, 0) & (4, 2, 0) & (5, 3, 0) & (4, 6, $\infty$) & (3, 4, $\infty$) & (100, 50, 0) & (200, 100, 0) & (250, 150, 0) \\
  \midrule
  GCN & 5.04 ± 0.13	&4.87 ± 0.17 &6.78 ± 0.19	&4.75 ± 0.10	&3.81 ± 0.16	&24.33 ± 0.53	&13.97 ± 0.52	&12.98 ± 0.45\\
  \cmidrule(lr){2-9}
  +IRM &2.42 ± 0.28  &3.21 ± 0.29 &3.08 ± 0.41 &2.39 ± 0.37 & 1.81 ± 1.11&2.72 ± 1.12 & 2.00 ± 0.90& 1.99 ± 0.70  \\
  +VREx &  5.10 ± 0.04 &5.60 ± 0.25 &6.91 ± 0.15 &4.67 ± 0.08 &3.92 ± 0.22 &23.06 ± 0.59 &12.53 ± 0.21 & 11.81 ± 0.31 \\
  +GroupDRO & 3.38 ± 0.26 &2.77 ± 0.07 &2.99 ± 0.02 &3.04 ± 0.75 &2.88 ± 0.74 &3.49 ± 0.64 &3.32 ± 0.42 & 3.49 ± 0.92 \\
  +DANN &5.47 ± 0.11   &5.40 ± 0.25 &6.98 ± 0.28 &8.63 ± 0.32 &6.65 ± 0.52 &21.83 ± 0.61 &12.53 ± 0.46 &11.67 ± 0.22  \\
  +Deep CORAL &5.13 ± 0.08  & 5.49 ± 0.13&6.82 ± 0.13 &8.47 ± 0.18 &6.72 ± 0.36 &23.10 ± 0.67 & 12.52 ± 0.20& 11.80 ± 0.31 \\
 \bottomrule
\end{tabular}
 \label{table:collab_ood_gen}
\end{adjustbox}
\end{table}

\begin{table}[h]
\centering
\caption{MRR Results for traditional OOD generalization methods on LPShift's {\bf ogbl-ppa}.}
\begin{adjustbox}{max width=\linewidth}
\begin{tabular}{l|ccc|cc|ccc}
\toprule
 \multirow{1}{*}{Models} & \multicolumn{3}{c}{CN Splits} &\multicolumn{2}{c}{SP Splits}  &\multicolumn{3}{c}{PA Splits}   \\ 
& (0, 1, 2) & (0, 2, 4) & (0, 3, 5) & ($\infty$, 6, 4) & ($\infty$, 4, 3) & (0, 5k, 10k) & (0, 10k, 20k) & (0, 15k, 25k) \\
  \midrule
  GCN & 7.81 ± 0.08&	8.18 ± 0.12&	8.75 ± 0.19&	6.34 ± 0.72	&5.05 ± 0.08	&3.69 ± 0.76	&3.42 ± 0.59	&5.62 ± 0.40\\
  \cmidrule(lr){2-9}
  +IRM &1.99 ± 0.45 &1.97 ± 1.37 &2.30 ± 1.05 &6.44 ± 0.62 &4.98 ± 0.15 &3.01 ± 0.61 &5.24 ± 0.20 &3.13 ± 0.56  \\
  +VREx &6.91 ± 0.18 &7.32 ± 0.25 &7.81 ± 0.21 &6.35 ± 0.83 & 5.19 ± 0.23&4.30 ± 0.77 &3.94 ± 0.83 & 4.71 ± 0.85 \\
  +GroupDRO &2.34 ± 0.19  &2.31 ± 0.27 &2.52 ± 0.37 &5.03 ± 1.91 & 2.62 ± 1.03&3.06 ± 0.44 &4.70 ± 0.60 &  2.77 ± 0.32 \\
  +DANN &7.06 ± 0.19  &7.34 ± 0.14 &7.82 ± 0.18 &6.34 ± 1.05 &5.30 ± 0.38 & 4.28 ± 0.87&4.74 ± 1.12 & 5.66 ± 0.36  \\
  +Deep CORAL & 6.91 ± 0.18 &7.32 ± 0.25 &7.79 ± 0.22 &6.25 ± 0.62 &5.18 ± 0.23 &4.19 ± 0.76 &4.08 ± 0.73 &  4.57 ± 0.74 \\
 \midrule
& (2, 1, 0) & (4, 2, 0) & (5, 3, 0) & (4, 6, $\infty$) & (3, 4, $\infty$) & (10k, 5k, 0) & (20k, 10k, 0) & (25k, 15k, 0) \\
  \midrule
  GCN & 2.59 ± 0.08&	2.16 ± 0.05&	2.17 ± 0.04&	9.95 ± 0.52&	3.29 ± 0.10	&3.09 ± 0.15	&2.50 ± 0.23&	2.79 ± 0.28 \\
  \cmidrule(lr){2-9}
  +IRM & 1.78 ± 0.56 &3.05 ± 0.17 &3.00 ± 0.32 &2.34 ± 2.07 &1.88 ± 0.64 &3.98 ± 0.23 &3.21 ± 0.28 & 3.09 ± 0.12  \\
  +VREx &3.51 ± 0.28   & 2.91 ± 0.11&2.87 ± 0.17 & 9.42 ± 0.85 &3.48 ± 0.32 &3.41 ± 0.86 &2.50 ± 0.48 & 2.57 ± 0.49 \\
  +GroupDRO &1.85 ± 0.42  & 3.20 ± 0.23&3.23 ± 0.16 &2.19 ± 0.29 &2.09 ± 0.60 &4.42 ± 0.26 &3.37 ± 0.08 &3.07 ± 0.13  \\
  +DANN & 3.62 ± 0.29  &2.72 ± 0.18 &2.73 ± 0.11 &9.69 ± 0.46 &3.32 ± 0.07 &2.71 ± 0.25 &2.13 ± 0.18 & 2.20 ± 0.20 \\
  +Deep CORAL & 3.43 ± 0.34 &2.93 ± 0.11 &2.70 ± 0.08 &9.42 ± 0.85 &3.48 ± 0.32 &2.57 ± 0.45 &2.13 ± 0.36 & 2.14 ± 0.15 \\
 \bottomrule
\end{tabular}
 \label{table:ppa_ood_gen}
\end{adjustbox}
\end{table}
\pagebreak

\begin{table}[h]
\centering
\caption{MRR Results for traditional OOD generalization methods on LPShift's {\bf ogbl-ddi}.}
\begin{adjustbox}{max width =\linewidth}
\begin{tabular}{l|ccc|ccc}
\toprule
 \multirow{1}{*}{Models} & \multicolumn{3}{c}{CN Splits} &\multicolumn{3}{c}{PA Splits}   \\ 
& (0, 1, 2) & (0, 2, 4) & (0, 3, 5) & (0, 5k, 10k) & (0, 10k, 20k) & (0, 15k, 25k) \\
  \midrule
  GCN &5.34 ± 0.24&	4.88 ± 0.41	&3.48 ± 0.26& 2.61 ± 0.38&	1.96 ± 0.22&	1.91 ± 0.14\\
  \cmidrule(lr){2-7}
  +IRM &0.77 ± 0.16 &1.36 ± 0.46 &1.26 ± 0.19 &1.34 ± 0.43 &1.55 ± 0.57 & 1.90 ± 1.17 \\
  +VREx &0.64 ± 0.02 & 0.61 ± 0.02& 1.41 ± 0.43& 2.09 ± 0.29&1.28 ± 0.71 &1.13 ± 0.21  \\
  +GroupDRO & 0.64 ± 0.02 &1.89 ± 0.72 &0.77 ± 0.16 & 1.12 ± 0.14&0.84 ± 0.22 &  2.22 ± 1.11 \\
  +DANN &0.63 ± 0.02  & 0.98 ± 0.82& 1.44 ± 0.50&2.19 ± 0.16 &1.12 ± 0.54 & 1.19 ± 0.20  \\
  +Deep CORAL & 0.64 ± 0.02 &0.61 ± 0.02 &1.29 ± 0.30 &2.22 ± 0.44 &1.23 ± 0.72 &  1.16 ± 0.20 \\
 \midrule
& (2, 1, 0) & (4, 2, 0) & (5, 3, 0) & (10k, 5k, 0) & (20k, 10k, 0) & (25k, 15k, 0) \\
  \midrule
  GCN &  1.14 ± 0.20	&1.01 ± 0.20	&1.41 ± 0.32 &4.68 ± 1.19 &2.04 ± 0.77 &1.14 ± 0.39\\
  \cmidrule(lr){2-7}
  +IRM &4.32 ± 0.64  &4.03 ± 1.02 &4.94 ± 2.16 &5.22 ± 2.91 & 5.69 ± 2.68 & 6.74 ± 0.85  \\
  +VREx &4.08 ± 0.63   &3.63 ± 0.01 &4.14 ± 0.97 &3.67 ± 1.76  &2.61 ± 2.36 &6.27 ± 0.58  \\
  +GroupDRO &4.58 ± 1.33  &3.68 ± 0.31 & 3.62 ± 0.01 & 4.13 ± 3.55&3.05 ± 3.23 & 6.37 ± 0.03 \\
  +DANN &  4.91 ± 1.31 &3.66 ± 0.10 &4.78 ± 1.48 &4.11 ± 2.44  &2.22 ± 2.03 &6.29 ± 1.32  \\
  +Deep CORAL &4.08 ± 0.63  &3.63 ± 0.01 &4.06 ± 1.00 &3.51 ± 1.63 & 1.71 ± 1.71& 6.19 ± 0.59 \\
 \bottomrule
\end{tabular}
 \label{table:ddi_ood_gen}
\end{adjustbox}
\end{table}

\section{Earth Mover's Distance (EMD) Results} \label{sec:app_emd_results}
This section contains the EMD calculations for all dataset splits and a visualization for the EMD calculations on the "Forward" ogbl-ddi dataset split.

\begin{figure}[h]
    \centering
    \includegraphics[width=\linewidth]{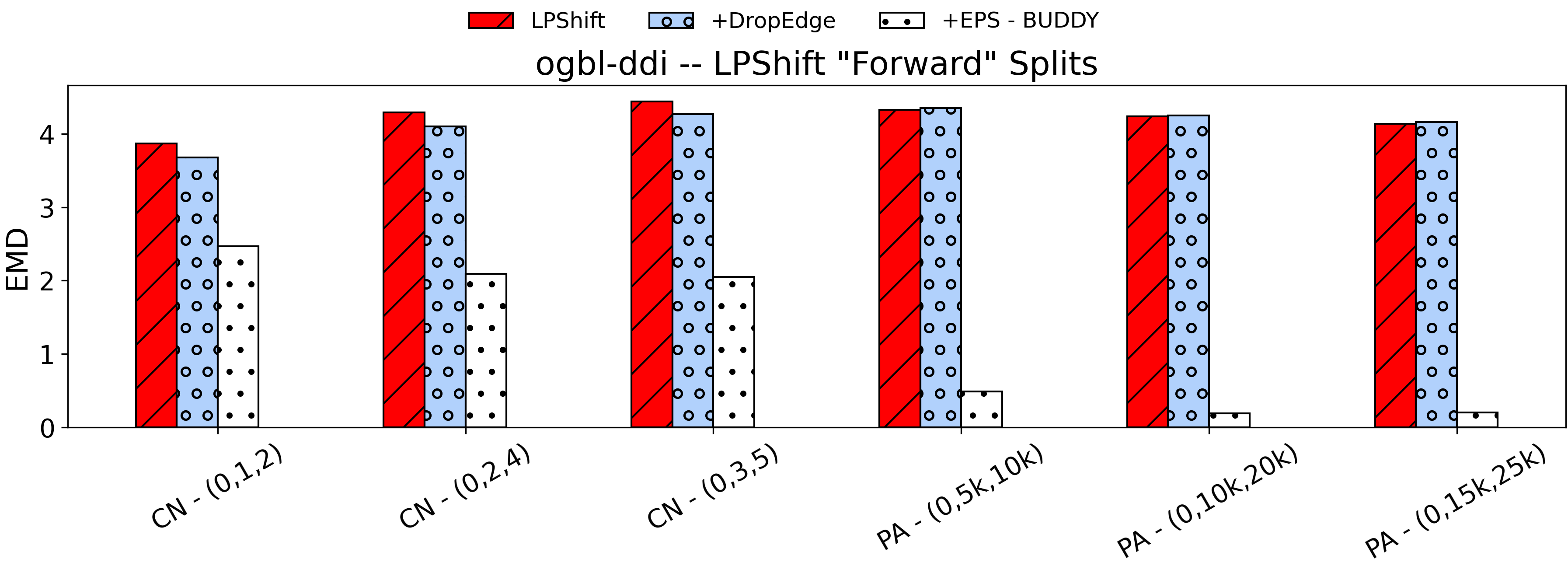}
    \caption{The EMD values calculated between the heuristic scores of training and testing samples on the "Backward" LPShift splits before and after applying structural generalization methods. \textit{Note:} The tested heuristics correspond to their labelled LPShift splits, so as to simulate the dataset splitting.}
    \Description{The EMD values calculated between the heuristic scores of training and testing samples on the "Backward" LPShift splits before and after applying structural generalization methods. \textit{Note:} The tested heuristics correspond to their labelled LPShift splits, so as to simulate the dataset splitting.}
    \label{fig:emd_ddi_plot}
\end{figure}

\begin{table}[h]
\centering
\caption{EMD calculations for ogbl-collab, ogbl-ppa, and ogbl-ddi on the forward and backward splits. (Note: Scores with a distance multiple-times different than the baseline are in \textbf{bold} for ogbl-collab.)}
\begin{adjustbox}{width=\linewidth}
\begin{tabular}{cc|ccc|ccc|ccc}
\toprule
\multicolumn{2}{c|}{Split} & \multicolumn{3}{c|}{ogbl-collab} & \multicolumn{3}{c|}{ogbl-ppa} & \multicolumn{3}{c}{ogbl-ddi} \\
\cmidrule(r){3-5} \cmidrule(l){6-8} \cmidrule(l){9-11}
 &  & Baseline & DropEdge & EPS & Baseline & DropEdge & EPS & Baseline & DropEdge & EPS \\
\midrule
\multirow{6}{*}{CN} 
  & (0, 1, 2) & 1.31 & 1.31 & \textbf{3.6}  & 2.82 & 2.82 & $>$24hrs & 3.87 & 3.68 & 2.47\\
  & (0, 2, 4) & 1.6  & 1.6  & 2.52         & 3.13 & 3.13 & $>$24hrs & 4.29 & 4.1 & 2.09\\
  & (0, 3, 5) & 1.45 & 1.65 & 2.22         & 3.05 & 3.19 & $>$24hrs & 4.44 & 4.27 & 2.05\\
\cmidrule{2-11}
  & (2, 1, 0) & 1.87 & 1.15 & 2.91         & 3.1  & 2.36 & $>$24hrs & 5.81 & 5.62 & 4.1\\
  & (4, 2, 0) & 2.26 & 1.49 & 2.99         & 3.3  & 2.55 & $>$24hrs & 5.81 & 5.62 & 4.17\\
  & (5, 3, 0) & 2.28 & 1.52 & 2.14         & 3.19 & 2.44 & $>$24hrs & 5.69 & 5.44 & 4.08\\
\midrule
\multirow{4}{*}{SP} 
  & ($\infty$, 6, 4) & 5.93 & 5.94 & \textbf{0.012} & 5.81 & 5.84 & $>$24hrs & -- & -- & -- \\
  & ($\infty$, 4, 3) & 5.35 & 5.38 & \textbf{0.003} & 1.36 & 1.4  & $>$24hrs & -- & -- & -- \\
\cmidrule{2-11}
  & (4, 6, $\infty$) & 3.6  & 3.53 & \textbf{1.22}  & 2.14 & 2.14 & $>$24hrs & -- & -- & -- \\
  & (3, 4, $\infty$) & 1.85 & 1.78 & 3.23         & 0.72 & 0.72 & $>$24hrs & -- & -- & -- \\
\midrule
\multirow{6}{*}{PA} 
  & (0, 50, 100)  & 1.87 & 1.89 & 3.42  & 2.55 & 2.55 & $>$24hrs & 4.33 & 4.35 & \textbf{0.49}\\
  & (0, 100, 200) & 2.29 & 2.32 & 2.72  & 2.76 & 2.76 & $>$24hrs & 4.24 & 4.25& \textbf{0.19}\\
  & (0, 150, 250) & 2.34 & 2.36 & 3.08  & 2.78 & 2.78 & $>$24hrs & 4.14 & 4.16 & \textbf{0.20}\\
\cmidrule{2-11}
  & (100, 50, 0)  & 4.29 & 4.3  & 2.48  & 2.96 & 2.96 & $>$24hrs & 6.85 & 6.92 & \textbf{0.71}\\
  & (200, 100, 0) & 3.79 & 3.82 & \textbf{0.78} & 2.68 & 2.68 & $>$24hrs & 6.41 & 6.49 & \textbf{0.58}\\
  & (250, 150, 0) & 3.48 & 3.5  & 2.48  & 2.48 & 2.48 & $>$24hrs & 5.98 & 6.02 & \textbf{0.44}\\
\bottomrule
\end{tabular}
\end{adjustbox}
\label{table:full_emd_dist}
\end{table}

\section{Dataset Licenses} \label{sec:app_data_licenses}

The dataset splitting strategy proposed in this paper is built using Pytorch Geometric (PyG). As such, this project's \href{https://github.com/revolins/LPShift}{software} and the \href{https://pytorch-geometric.readthedocs.io/en/latest/}{PyG datasets} are freely-available under the MIT license.

\section{Limitations} \label{sec:app_limitations}

The proposed dataset splitting strategy is restricted to inducing distribution shifts solely with neighborhood heuristics on static graphs. So, it does not directly consider other types of possible distribution shifts for the link prediction task (i.e. spatio-temporal \cite{zhang2022dynamic} or size \cite{zhou2022ood} shift). Additionally, since the neighborhood heuristics compute discrete scores produced from an input graph's structural information and effectively training GNN4LP models requires no leakage with validation/testing, it may be difficult to determine the correct thresholds to extract a meaningful number of samples. For Common Neighbors and Preferential-Attachment, this is especially relevant with smaller training graphs, given that larger and/or denser graphs have inherently more edges. Therefore, larger and denser graphs have inherently more possible Common Neighbors and Preferential-Attachment scores. For Shortest-Path, splitting can be exceptionally difficult for denser graphs, as demonstrated with the tiny split sizes for ogbl-ppa in Table~\ref{table:ppa_samp_size}.

\section{Impact Statement} \label{sec:app_impact}

Our proposed dataset splitting strategy mimics the formatting of PyTorch Geometric datasets. This means that our strategy is simple to implement, enabling future work involved with understanding this type of structural shift for link prediction and promoting beginner-friendly practices for artificial intelligence research. Additionally, since the structural shift we propose in this article affects real-life systems, which integrate link prediction models, this research can provide a foundation for the improvement of relevant technologies; which holds positive ramifications for society and future research. No apparent risk is related to the contribution of this work.

\end{document}